\documentclass[a4paper,fleqn]{cas-sc}

\usepackage{amssymb}
\usepackage{amsmath}
\usepackage{hyperref}
\usepackage{graphicx}
\usepackage{amssymb}
\usepackage{pifont}
\usepackage{array}
\usepackage{float}
\usepackage{makecell}
\usepackage{multirow}
\usepackage{longtable}
\usepackage{supertabular}
\usepackage{placeins}
\usepackage{natbib}
\usepackage{mathtools}
\usepackage{amsmath}
\setcitestyle{numbers}
\usepackage{soul}
\usepackage{xcolor}
\usepackage{ulem}

\begin{document}

\let\WriteBookmarks\relax
\def\floatpagepagefraction{1}
\def\textpagefraction{.001}



\title[mode = title]{ShaTS: A Shapley-based Explainability Method for Time Series Artificial Intelligence Models applied to Anomaly Detection in Industrial Internet of Things}

\author[1]{Manuel Franco de la Peña}[orcid=0009-0008-8583-5942]

\ead{manuel.francop@um.es}
\affiliation[1]{organization={Departamento de Ingenier\'ia y Tecnolog\'ia de Computadores, University of Murcia}, addressline={Espinardo}, city={Murcia}, country={Spain} }

\author[1]{\'Angel Luis Perales G\'omez}[orcid=0000-0003-1004-881X]\cormark[1]
\ead{angelluis.perales@um.es} \cortext[cor1]{Corresponding author}



\author[1]{Lorenzo Fern\'andez Maim\'o}[orcid=0000-0003-2027-4239]
\ead{lfmaimo@um.es}

\begin{abstract}
Industrial Internet of Things environments increasingly rely on advanced Anomaly Detection and explanation techniques to rapidly detect and mitigate cyberincidents, thereby ensuring operational safety. The sequential nature of data collected from these environments has enabled improvements in Anomaly Detection using Machine Learning and Deep Learning models by processing time windows rather than treating the data as tabular. However, conventional explanation methods often neglect this temporal structure, leading to imprecise or less actionable explanations. This work presents ShaTS (Shapley values for Time Series models), which is a model-agnostic explainable Artificial Intelligence method designed to enhance the precision of Shapley value explanations for time series models. ShaTS addresses the shortcomings of traditional approaches by incorporating an a priori feature grouping strategy that preserves temporal dependencies and produces both coherent and actionable insights. Experiments conducted on the SWaT dataset demonstrate that ShaTS accurately identifies critical time instants, precisely pinpoints the sensors, actuators, and processes affected by anomalies, and outperforms SHAP in terms of both explainability and resource efficiency, fulfilling the real-time requirements of industrial environments.
\end{abstract}










\date{}
\maketitle

\section{Introduction}
\label{sec1}

Industry 5.0 has fundamentally transformed industrial environments by enhancing human-machine collaboration, connectivity, and data-driven decision-making \cite{industria5_0}. A key component of this transformation is the deployment of interconnected sensors and actuators within Industrial Internet of Things (IIoT) systems, which continuously generate time-series data. This data provides a detailed representation of industrial process dynamics and enables the application of Big Data \cite{bigdata_empresa} and Artificial Intelligence \cite{ia_empresa} methods.

This increased connectivity exposes industrial facilities to heightened cyberattack risks, rendering conventional Intrusion Detection Systems (IDS) increasingly obsolete  \cite{empresa_conectada, empresa_ataques}. Consequently, Anomaly Detection (AD) techniques have emerged as a more effective alternative. The AD models predominantly employ Machine Learning (ML) and Deep Learning (DL) algorithms that exploit the sequential nature of time-series data to learn normal operational patterns and accurately detect deviations indicative of cyberattacks or system malfunctions. However, despite their strong performance, these models typically function as black boxes, making it difficult to quickly identify root causes and implement targeted interventions, a significant drawback in industrial settings where timely responses are essential.

Explainable Artificial Intelligence (xAI) aims to overcome the opacity of these complex ML/DL models \cite{molnar2022}. Among the most widely applied methods in xAI are those based on Shapley values, which provide a systematic approach to quantifying the contribution of each input feature to a model's prediction. However, despite their popularity, conventional Shapley-based methods often overlook the temporal dependencies intrinsic to time-series data, leading to explanations that may be incomplete or even misleading. Additionally, these methods typically incur significant computational overhead when calculating Shapley values, which can hinder real-time deployment in IIoT settings, as delays in these environments may compromise safety and even endanger personnel. They also lack inherent mechanisms for aggregating features that represent the same underlying physical component or process, leading to fragmented and less actionable insights. Addressing these limitations is essential to ensure that the insights derived from AD models are accurate and actionable, ultimately enhancing the reliability and operational effectiveness of industrial cybersecurity measures.

To address these limitations, this work introduces \textbf{Shapley values for Time Series models (ShaTS)}, a Shapley-based xAI method specifically designed for time-series data with applicability to AD frameworks in IIoT. The main contributions of this work are as follows:

\begin{itemize}
    \item \textbf{Open-Source xAI Module}: ShaTS is a fully documented, plug-and-play xAI module designed for seamless integration into ML/DL frameworks that deal with time-series data. While broadly applicable across various domains, it is particularly tailored to improve AD in IIoT. The module is released as open source and is available on GitHub \footnote{\url{https://github.com/CyberDataLab/ShaTS}}.

    \item \textbf{Feature Grouping for Enhanced Explainability}: ShaTS introduces a novel approach to feature grouping, applied prior to computing feature importance, to improve the explainability of time-series data ML/DL models and deliver more contextually relevant explanations. This methodology is structured into three distinct strategies: Temporal Grouping, Feature Grouping, and Multi-Feature Grouping, each designed to capture different aspects of the data's influence on model predictions.

    \item \textbf{Validation on the SWaT Dataset}: ShaTS is evaluated using the Secure Water Treatment (SWaT) dataset, widely used for AD in IIoT environments. Experimental results confirm that ShaTS accurately identifies the critical time instants that drive anomaly predictions and effectively locates the specific sensors/actuators and processes responsible for the anomalies. Furthermore, a comparative analysis with the widely used SHAP library demonstrates that ShaTS’s a priori grouping strategy better preserves temporal dependencies, resulting in more precise and actionable insights for industrial cybersecurity. Additionally, ShaTS demonstrates lower computational resource consumption and faster execution times, thereby meeting the stringent real-time requirements of modern IIoT environments.

\end{itemize}

The paper is structured as follows: Section \ref{sec_marco} provides the theoretical basis, Section \ref{sec:relatedWork} reviews relevant literature, Section \ref{sec_proposed} presents the proposed xAI module, Section \ref{sec_experimentation} details experimental results, and Section \ref{sec_conclusion} discusses conclusions and future work.

\section{Theoretical basis}
\label{sec_marco}

Within the realm of xAI, quantifying the contribution of each feature to a model's prediction is crucial for enhancing transparency and understanding the underlying decision-making processes. Cooperative game theory offers a foundational approach to this challenge by defining methods that assess the impact of individual \textit{players} within a collaborative setting. Among these methods, the Shapley value stands out as a technique that fairly allocates a contribution to each participant based on their impact on the final outcome. By treating each input feature as a \textit{player} whose contribution for a model's prediction is to be assessed, this concept can be easily adapted to serve as a xAI technique.

 In cooperative game theory, when a set of players \( N \) collaborates to achieve a common objective, the Shapley value assesses each player’s individual contribution to that outcome. A cooperative game is represented as a function \( v: 2^N \to \mathbb{R} \), which assigns a real value to each possible coalition of players \( S \subset N \). The Shapley value (Equation \ref{eq:shapley1953}) for player \( i \), denoted \( \varphi_i(v) \), is calculated by averaging the marginal contributions of player \( i \) across all possible coalitions \( S \) that do not include \( i \), weighted by the probability of the corresponding coalition forming in a random order.

\begin{small}
\begin{equation}
    \varphi_i(v) = \sum_{S \subseteq N \setminus \{i\}} \frac{|S|!(|N|-|S|-1)!}{|N|!} \cdot \left[ v(S \cup \{i\}) - v(S) \right]
\label{eq:shapley1953}
\end{equation}
\end{small}

Rearranging the terms of Equation \ref{eq:shapley1953}, the Shapley value $\varphi_i(v)$ of a player $i$ can also be expressed as Equation \ref{eq:shapleyAlternativo1}, where the exact value of every term $\varphi_i^j(v)$ is detailed in Equation \ref{eq:shapleyAlternativo2} and $S_i^j$ is defined as the set of all possible coalitions of size $j$ that do not include player $i$.

\begin{small}
\begin{equation}
    \varphi_i(v) = \frac{1}{|N|} \sum_{j=0}^{|N|-1} \varphi_i^j(v)
\label{eq:shapleyAlternativo1}
\end{equation}
\end{small}
\begin{small}
\begin{equation}
    \varphi_i^j(v) = \frac{1}{|S^j_i|} \sum_{S \in S^j_i}  \cdot \left[ v(S \cup \{i\}) - v(S) \right]
\label{eq:shapleyAlternativo2}
\end{equation}
\end{small}

The Shapley value satisfies four axioms that make it a fair and unique solution for distributing value in cooperative settings. These axioms are as follows:

\begin{itemize}
    \item \textbf{Symmetry}: If two players \( i \) and \( j \) contribute equally to every coalition that excludes both of them, they must receive the same Shapley value. Formally, if for all subsets \( S \subseteq N \setminus \{i, j\} \), it holds that \( v(S \cup \{i\}) = v(S \cup \{j\}) \), then \( \varphi_i(v) = \varphi_j(v) \).

    \item \textbf{Additivity}: If two cooperative games, \( v \) and \( w \), are considered, then for each player \( i \), the Shapley value of the combined game \( v + w \) must equal the sum of the Shapley values from each game independently: \( \varphi_i(v + w) = \varphi_i(v) + \varphi_i(w) \). 

    \item \textbf{Efficiency}: The sum of Shapley values across all players must equal the total value of the grand coalition, where all players participate. This is mathematically expressed as \( \sum_{i \in N} \varphi_i(v) = v(N) \). 

    \item \textbf{Dummy Player (Null Player)}: If a player \( i \) does not change the value of any coalition beyond their own individual value, it must receive a Shapley value equal to this individual contribution. Formally, if for every subset \( S \subseteq N \) that does not contain \( i \), it holds that \( v(S \cup \{i\}) = v(S) + v(\{i\}) \), then \( \varphi_i(v) = v(\{i\}) \). 
\end{itemize}

Building on these theoretical foundations, the Shapley value has been adapted for use in xAI as a means to determine the contribution of each input feature for every model's prediction. For any subset of features $S$, let $x$ denote the complete input feature vector. Then, $x^S$ represents the values corresponding to the features in $S$, while $x^{\overline{S}}$ corresponds to those not in $S$. The model's conditional value for $S$ in a specific instance $x_*$ is defined as $v(S, x_*) = \mathbb{E}(f(x) \mid x^S = x_*^S)$ where $f(\cdot)$ is the model's prediction function. Using this definition, the Shapley value for each feature quantifies its contribution to the prediction and is computed as shown in Equation \ref{eq:shapleyML}.

\begin{footnotesize}
\begin{equation}
\varphi_i(v, x_*) = \sum_{S \subseteq N \setminus \{ i \}} \frac{|S|! (|N| - |S| - 1)!}{|N|!} \cdot \left[ v(S \cup \{ i \}, x_*) - v(S, x_*) \right]
\label{eq:shapleyML}
\end{equation}
\end{footnotesize}

\section{Related work} \label{sec:relatedWork}

This section reviews state-of-the-art xAI approaches for AD in IIoT, with a particular focus on methods that leverage Shapley values and address the temporal dependencies inherent in time-series data. The review is organized into two parts: one covering Shapley-based xAI techniques and another discussing broader xAI approaches applied to AD in IIoT.

\subsection{Shapley-based xAI methods}

Shapley-based methods have emerged as a cornerstone in xAI for attributing the contribution of individual features to a model’s prediction. However, their practical application is hindered by two primary challenges \cite{Chen2023}.

The first challenge stems from the need to estimate the conditional value function, defined as $v(S, x_*) = \mathbb{E}(f(x) | x^S = x_*^S)$, which represents the expected output of the model $f(\cdot)$ when only the features in the subset $S$ are provided with the values observed in the instance $x_*$. Since most ML/DL models require complete input vectors, the values for features not included in $S$ must be approximated or imputed. A common approach \cite{baseline1, baseline2} is to replace the missing features with predetermined baseline values (e.g., the mean or zero); however, this method can introduce bias if the chosen baselines do not accurately reflect the underlying data distribution. Alternatively, other methods \cite{kernelshap}, assume feature independence and approximate the conditional expectation using marginal distributions, i.e., $v(S, x_*) \approx \mathbb{E}(f(x) | x^S)$. More advanced approaches seek to estimate the full conditional distribution without assuming independence, either by adopting parametric models (e.g., assuming a Gaussian distribution) or by employing empirical estimations that weight training instances according to their similarity to $x_*$ \cite{improvingkernelshap}. Typically, these estimations are performed on a representative subset of the training dataset, the background dataset, to balance computational feasibility with estimation accuracy.

The second challenge arises from the computational complexity inherent in Shapley value calculation, primarily due to the exponential growth in the number of feature subsets that must be evaluated. To mitigate this issue, various approximation strategies have been proposed, which can be broadly classified into model-dependent and model-independent methods. Model-dependent approaches exploit specific properties of the underlying model to streamline the computation. For example, TreeSHAP \cite{treeSHAP} leverages the inherent structure of decision tree ensembles to compute exact Shapley values in polynomial time. Similarly, linearSHAP and DeepSHAP \cite{kernelshap} are tailored to linear models and deep neural networks, respectively. Conversely, model-independent methods do not rely on the internal structure of the predictive model. A common strategy within this class involves approximating the local behavior of the model by fitting a surrogate model, often formulated as a least squares problem, in the vicinity of the instance to be explained. Methods such as fastSHAP \cite{fastShap}, KernelSHAP \cite{kernelshap}, and sgdSHapley \cite{sgdShap} adopt this approach, with the advantage that a well-fitted surrogate can yield closed-form solutions for Shapley values in certain cases. Alternatively, some methods approximate the Shapley summation directly by sampling only a subset of coalitions, either based on their likelihood of occurrence or via random sampling followed by appropriate weighting \cite{approshap}. Further alternatives include adapting the predictive model to handle missing features during training or training a surrogate DL model specifically designed to approximate coalition values \cite{surrogate}.

In conclusion, while challenges in conditional value estimation and computational complexity persist, effective strategies have been developed to render Shapley-based methods practicable. The widespread adoption of these techniques is evidenced by the prominence of the SHAP repository \cite{kernelshap}, which has become the standard toolkit for feature attribution in xAI.

\subsection{xAI modules in AD IIoT frameworks}

Recent advances in ML/DL have significantly improved AD performance in IIoT by leveraging the temporal dynamics inherent in sensors and actuators measurements. Nevertheless, the black-box nature of these models, combined with the criticality of IIoT environments, has heightened the need for xAI \cite{survey_ad_ics}. Many existing xAI methods, however, are designed for tabular data and compute feature importance independently at each time instant. This practice neglects the sequential behavior of time-series data and can result in explanations that are neither accurate nor actionable, an outcome that undermines the primary objective of xAI, which is to enable precise and effective interventions.

\begin{table*}[htbp]
\centering
\footnotesize
\begin{tabular}{l||>{\centering\arraybackslash}m{1.8cm}|>{\centering\arraybackslash}m{1.8cm}||>{\centering\arraybackslash}m{1.8cm}|>{\centering\arraybackslash}m{1.8cm}|>{\centering\arraybackslash}m{1.8cm}|c|c}
\hline
\multicolumn{1}{c||}{\textbf{Sol.}} & \multicolumn{2}{c||}{\textbf{AD Module}} & \multicolumn{5}{c}{\textbf{xAI Module}} \\ \hline
\textbf{} & \textbf{Model} & \makecell{\textbf{Temporal}\\ \textbf{AD}} & \textbf{Method} & \makecell{\textbf{Agnostic}} & \makecell{\textbf{Feature}\\ \textbf{Grouping}} & \makecell{\textbf{Actionable}\\ \textbf{insights}} & \makecell{\textbf{Temporal}\\ \textbf{xAI}} \\ \hline

\cite{brito2022explainable} & \makecell{Isolation Forest\\ kNN} & \ding{51} & \makecell{SHAP} & \ding{51} & None & \ding{55} & \ding{55} \\ \hline

\cite{3-huong2022federated} & \makecell{VAE\\ SVDD} & \ding{51} & SHAP & \ding{51} & None & \ding{55} & \ding{55} \\ \hline

\cite{brusa2023explainable} & SVM & \ding{51} & SHAP & \ding{51} & None & \ding{55} & \ding{55} \\ \hline

\cite{6-jacob2020exathlon} & \makecell{LSTM\\AE} & \ding{51} & \makecell{LIME\\ MacroBASE\\ EXstream} & \ding{51} & None & \ding{55} & \ding{55} \\ \hline

\cite{hong2020} & CNN-LSTM & \ding{51} & SHAP & \ding{51} & \makecell{Post hoc} & \ding{51} & \ding{55} \\ \hline

\cite{4-mathuros2024waxai} & \makecell{ECOD\\ DeepSVDD} & \ding{51} & \makecell{SHAP\\ LIME\\ ALE} & \ding{51} & \makecell{Post hoc} & \ding{51} & \ding{55} \\ \hline

\cite{9-fung2024attributions} & \makecell{LSTM \\ GRU \\ CNN} & \ding{51} & \makecell{SHAP\\ LIME\\ LEMNA} & \ding{51} & \makecell{Post hoc} & \ding{51} & \ding{55} \\ \hline

\cite{1-hoang2022explainable} & LSTM-AE & \ding{51} & SHAP & \ding{51} & \makecell{Post hoc} & \ding{51} & \ding{55} \\ \hline

\cite{2-hwang2021sfd} & Bi-LSTM & \ding{51} & SHAP & \ding{51} & \makecell{Post hoc} & \ding{51} & \ding{55} \\ \hline

\cite{11-tang2023gru} & \makecell{GRU\\GNN} & \ding{51} & \makecell{GRN\\t-SNE} & \ding{55} & \makecell{Post hoc} & \ding{51} & \ding{55} \\ \hline

\cite{10-perales2023interpretable} & LSTM & \ding{51} & Causal Inference & \ding{55} & \makecell{A priori} & \ding{51} & \ding{55} \\ \hline

Ours & LSTM & \ding{51} & ShaTS & \ding{51} & \makecell{A priori} & \ding{51} & \ding{51} \\ \hline
\end{tabular}
\caption{Comparison of xAI Approaches in AD for IIoT. The table summarizes each framework’s AD module (including the model used and whether it employs temporal analysis), along with details of its xAI module: the technique employed, whether it is model-agnostic, the feature grouping strategy (post hoc or a priori), the provision of actionable insights, and the incorporation of temporal considerations in the xAI process.}
\label{table:state}
\end{table*}

Several frameworks integrate xAI modules into their AD systems for IIoT; however, many of these approaches do not yield actionable explanations. For instance, \cite{brito2022explainable} employs SHAP in an unsupervised AD framework for rotating machinery using models such as Isolation Forest (IF) and k-Nearest Neighbors (kNN). This approach uses a 100-second time window to extract statistical features, such as kurtosis and root mean square, from each sensor, yet computes SHAP values independently for each feature without aggregating them at the sensor level, thereby limiting the practical utility of the explanations. Similarly, \cite{3-huong2022federated} presents FedeX, a distributed detection architecture that combines a Variational Autoencoder (VAE) with Support Vector Data Description (SVDD). Although features are extracted from sensor data and SHAP is applied to each feature, no mechanism is provided to consolidate these values into sensor-level attributions. In another study, \cite{brusa2023explainable} applies SHAP to supervised models like Support Vector Machine (SVM) for fault diagnosis in IIoT; while this approach enhances transparency, it fails to deliver actionable insights because it computes SHAP values for individual features, such as root mean square, skewness, and shape factor, in isolation. Finally, the Exathlon benchmark \cite{6-jacob2020exathlon} apply xAI techniques such as Local Interpretable Model-Agnostic Explanations (LIME) and EXstream to AD models like LSTM (Long Short-Term Memory), MacroBASE and an AE (AutoEncoder). However, this framework does not generate explanations that facilitate targeted interventions.

Some frameworks, in contrast, do yield actionable explanations by tracing anomalies to the specific sensors or actuators involved. These approaches typically compute feature-level attributions and then aggregate them, usually by summing the importance scores of all features associated with a given sensor, to derive sensor-level insights. For example, \cite{hong2020} employs SHAP with a Convolutional Neural Network (CNN)-LSTM model for predicting the remaining useful life of turbofan engines. In this framework, the model is trained on raw sensor measurements collected over a time window, and sensor-level importance is obtained by summing Shapley values calculated at each time instant. Similarly, \cite{4-mathuros2024waxai} investigates interpretability techniques in an AD context using Empirical Cumulative distribution based Outlier Detection (ECOD) and Deep Support Vector Data Description (DeepSVDD), applying methods such as SHAP, LIME, and Accumulated Local Effects (ALE); these methods, too, perform grouping post hoc by aggregating feature attributions. Authors in \cite{9-fung2024attributions} also adopts a strategy in which sensor importance is determined by summing feature attributions computed across time. In their study, the AD models tested include CNNs, Gated Recurrent Unit Networks (GRUs), and LSTMs, while the xAI methods employed comprise LIME, SHAP, and the Local Explanation Method using Nonlinear Approximation (LEMNA). In a similar vein, both \cite{1-hoang2022explainable} and \cite{2-hwang2021sfd} leverage SHAP within LSTM-based frameworks, an LSTM-AE and a Bidirectional-LSTM (Bi-LSTM) respectively, aggregating individual feature contributions to obtain sensor-level insights.  Although these methods improve actionability by identifying the sensor or actuator driving the anomaly, the reliance on post hoc aggregation may still inadequately capture the inherent temporal dependencies, highlighting the need for more integrated approaches.

Finally, alternative strategies have been proposed that yield actionable explanations without relying solely on the summation of individual feature importances. For instance, \cite{10-perales2023interpretable} employs a temporal window approach in which features are grouped by sensor and an unsupervised model is used to flag anomalies where if any of the grouped features exceed a predefined threshold, the corresponding sensor is identified as the source of the anomaly. Moreover, this study also explores a causal inference approach within an LSTM model that distinguishes between cause and effect in anomalies by performing feature grouping a priori, thereby further enhancing interpretability. Similarly, \cite{11-tang2023gru} introduces a novel method that integrates Graph Neural Networks (GNNs) and GRUs with t-SNE to identify and explain anomalies. In this approach, feature importance is quantified by computing a normalized measure of the maximum discrepancy between the predicted and actual values, rather than simply aggregating individual contributions. However, both approaches depend on a predetermined threshold for AD, which can limit their flexibility and application.

In summary, these limitations motivate the proposed module, Shapley values for Time Series models (ShaTS). Unlike existing Shapley-based methods that rely on post hoc aggregation, ShaTS integrates a priori feature grouping directly into the computation of Shapley values. Moreover, it offers a range of complementary grouping strategies tailored to different levels of analysis, depending on the explanatory needs. This flexibility not only preserves the temporal dependencies inherent in sensor data but also provides actionable explanations that support more precise and effective interventions in critical IIoT environments.

Table \ref{table:state} summarizes the reviewed AD frameworks by detailing their AD modules, the xAI techniques employed, and the grouping strategies implemented. This comparison distinguishes between post hoc grouping, where individual feature attributions are aggregated after computation, and a priori grouping, where features are grouped as part of the attribution process, and also highlights whether the methods produce actionable insights and incorporate temporal xAI considerations.

\section{Proposed xAI Module: Integrating ShaTS into an AD Framework for IIoT}
\label{sec_proposed}

\begin{figure*}
    \centering
    \includegraphics[width=\linewidth]{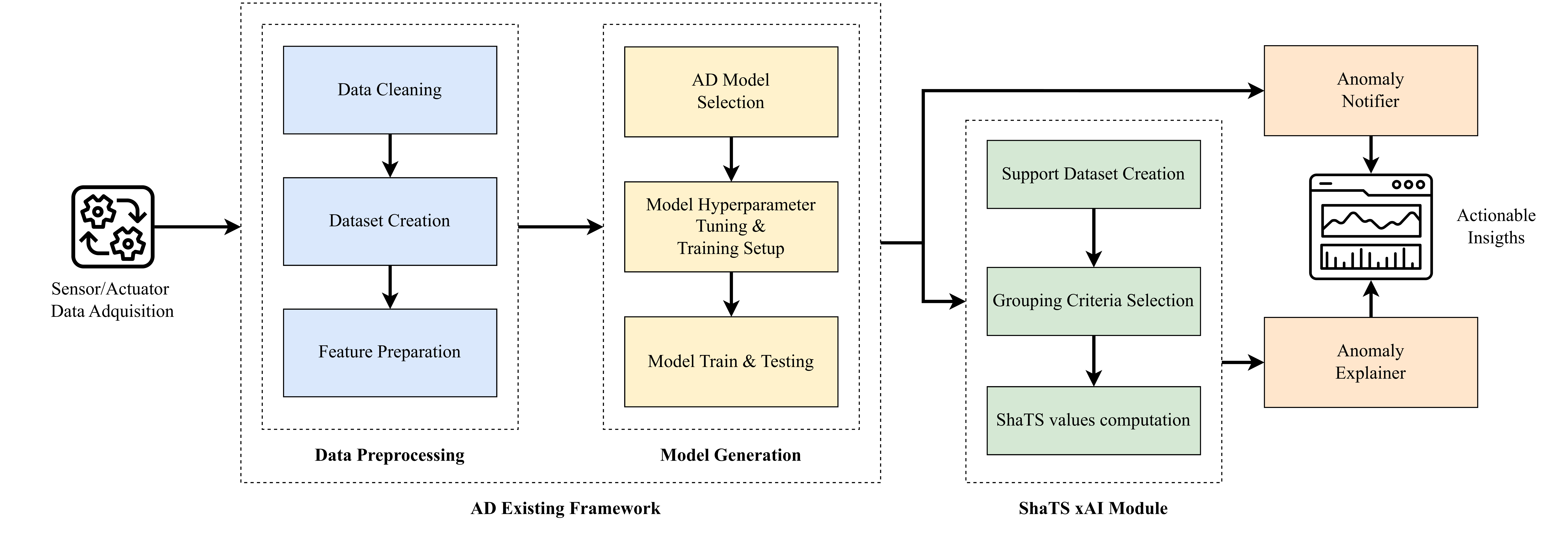}
    \caption{ShaTS xAI Module integration in a typical AD framework}
    \label{fig:proposedFramework}
\end{figure*}

In this section, the xAI module ShaTS is introduced as a flexible component designed to enhance explainability in ML/DL models that make use of time-series data. While ShaTS is a domain-agnostic module, its implementation is demonstrated within an AD framework for IIoT as a practical use case. A typical framework for this purpose consists of two core components: a Data Preprocessing Module, where data is prepared for the ML/DL model, and a Model Generation Module, responsible for selecting and training an AD model. The integration of the ShaTS xAI module into a typical AD framework is illustrated in Figure \ref{fig:proposedFramework}.

\subsection{Data Preprocessing Module}

The Data Preprocessing Module is designed to prepare IIoT data effectively for AD model training. This process is organized into three main components: Data Cleaning, Dataset Creation, and Feature Preparation.

\subsubsection{Data Cleaning}

Data cleaning is essential to remove noise and irrelevant entries that may obscure the detection of meaningful patterns. Noise in IIoT data often arises from sources such as sensor drift, intermittent failures, or unexpected environmental interferences. To address these issues, noise is reduced using filters like moving averages or low-pass filters, which smooth out erratic fluctuations while retaining significant patterns. Additionally, missing values are handled through imputation or selective exclusion of incomplete data points, depending on the frequency and distribution of these gaps.

\subsubsection{Dataset Creation}

\begin{figure}[htbp]
    \centering
    \includegraphics[width=0.6\linewidth]{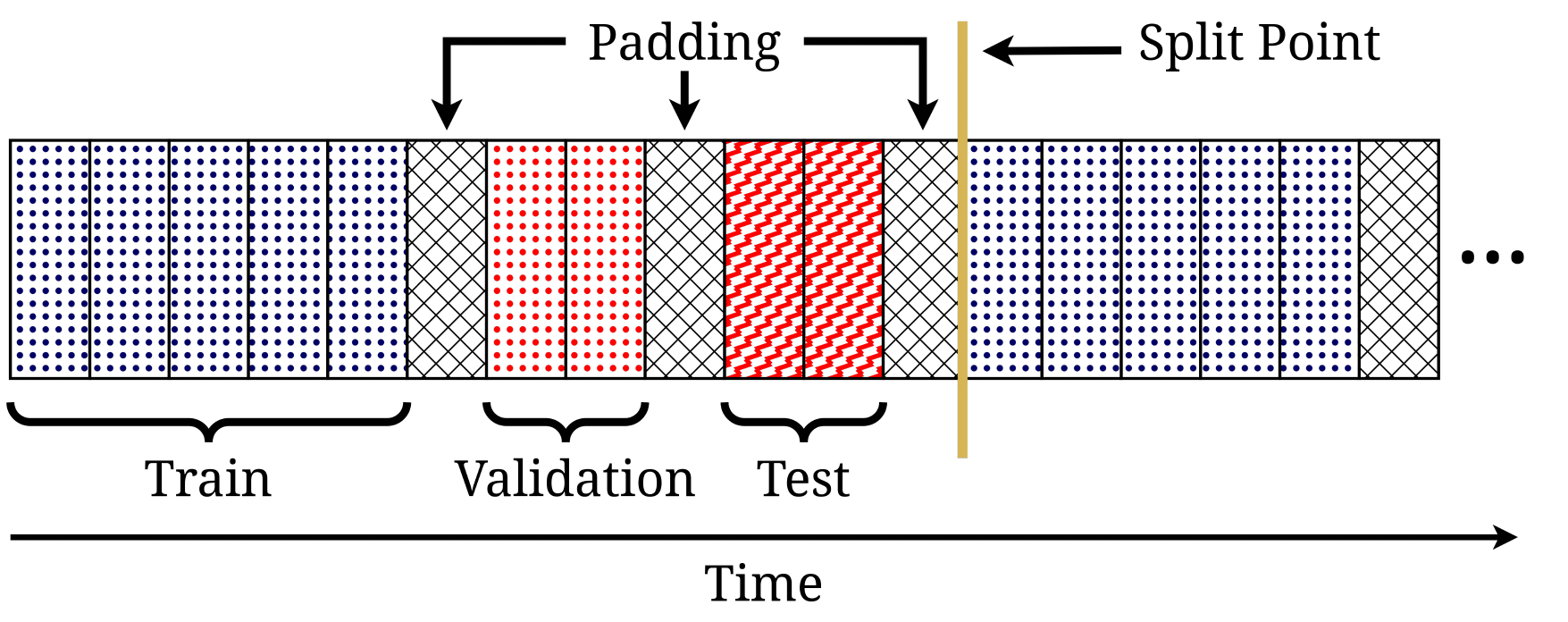}
    \caption{Padding in the dataset partitiong process to prevent data leakage between segments.}
    \label{fig:padding}
\end{figure}

After Data Cleaning, the following step is the creation of training, validation, and test datasets.  Given the temporal nature of IIoT data, it is crucial to preserve temporal dependencies within these datasets. To ensure this, the data is divided at predefined split points, creating distinct segments. Each segment is then divided sequentially into training, validation, and test subsets, with padding applied between them to prevent data leakage. Finally, the complete training, validation, and test datasets are formed by concatenating these subsets from each segment (see Figure \ref{fig:padding}).

Within each segment, data is further organized into overlapping time windows to capture temporal patterns. Each window encapsulates a complete snapshot of sensor and actuator readings over a defined period. The window size defines the duration of each snapshot, while the stride specifies the time interval between the start of consecutive windows. These parameters are carefully chosen to balance redundancy with the need for sufficient temporal coverage, ensuring that both long-term and short-term patterns are represented without excessive repetition.

\subsubsection{Feature Preparation}

The feature preparation step involves organizing sensor and actuator data to optimize the model's ability to distinguish between normal and anomalous patterns. Many IIoT datasets include categorical variables, such as the states of valves or pumps, which are converted into binary vectors using one-hot encoding. This ensures that categorical variables are properly represented for effective processing by the ML/DL model.

To manage the large number of features effectively, a redundancy reduction process is also performed, identifying and removing features with zero variance, as these provide no useful information for the model. 

Finally, all continuous features are normalized to ensure that no single feature disproportionately influences model outcomes. Standard normalization adjusts features to have a mean of zero and a standard deviation of one, centering and scaling the data for consistent distribution. Alternatively, min-max normalization scales features to a fixed range, typically between zero and one, preserving relative differences while bounding values within a defined range.

\subsection{Model Generation Module}

The Model Generation module is responsible for designing and training the AD model. This module is divided into three main components: AD Model Selection, Model Hyperparameter Tuning \& Training Setup, and Model Train \& Testing. 

\subsubsection{AD Model Selection}

Selecting an appropriate AD model is guided by the characteristics of the IIoT data, the application requirements, and the desired format of the model’s output. One key factor in model selection is the availability of labeled data. When labeled data is available, supervised models are preferred as they can explicitly differentiate normal from anomalous states, which enhances detection accuracy. In the absence of labeled data, unsupervised models, such as autoencoders or clustering-based methods, are used to detect anomalies based solely on inherent patterns in the data.

Another consideration is the model's ability to handle time-series data, given the temporal nature of IIoT datasets. Models like LSTM networks, RNNs, and CNNs are commonly chosen for this purpose. LSTM networks are particularly well-suited for capturing long-term dependencies, while CNNs can efficiently detect localized patterns within time windows.

Finally, ML models can produce outputs in various formats, but AD models are typically designed to yield a probability score that reflects the likelihood of an anomaly. This score is often derived by applying a threshold to the model’s output, where values above the threshold indicate a high anomaly probability. In other cases, specific activation functions are used to adjust the model’s output to reflect this probability directly.

\subsubsection{Model Hyperparameter Tuning \& Training Setup}

This step involves hyperparameter tuning and loss function selection to optimize performance. During hyperparameter tuning, different configurations, such as learning rate, batch size, and network structure, are explored using strategies like grid search or random search. Grid search systematically evaluates all possible combinations within a defined range, providing thorough but computationally intensive coverage of the hyperparameter space. Random search, on the other hand, explores a randomized subset of configurations, which can be computationally efficient in cases where only a few hyperparameters have a significant impact on performance. The validation dataset plays a critical role in assessing the performance of each configuration, ensuring the selected hyperparameters lead to a model that generalizes effectively to unseen data.

The loss function is then chosen based on the specific challenges of AD. Binary cross-entropy (Equation \ref{eq:crossentropy}) is commonly used for balanced binary classification. However, in cases with severe class imbalance, as often encountered in AD, Focal Loss (Equation \ref{eq:focalloss}) is preferred. Focal Loss enhances model sensitivity by focusing on hard-to-classify samples and down-weighting the contribution of well-classified instances. This loss function includes hyperparameters $\alpha$ for balancing class weights and $\gamma$ for focusing on challenging samples, allowing the loss function to adapt to the imbalanced data distribution.

    \begin{small}
    \begin{equation} 
    L(y, \hat{y}) = -\frac{1}{N} \sum_{i=1}^{N} \left[ y_i \log(\hat{y}_i) + (1 - y_i) \log(1 - \hat{y}_i) \right] 
    \label{eq:crossentropy}
    \end{equation}
    \end{small}

\begin{small}
\begin{equation}
\begin{aligned}
L_{\text{focal}}(y, \hat{y}) = -\frac{1}{N} \sum_{i=1}^{N} \Big[ & \, \alpha y_i (1 - \hat{y}_i)^{\gamma} \log(\hat{y}_i) + (1 - \alpha)(1 - y_i) \hat{y}_i^{\gamma} \log(1 - \hat{y}_i) \Big]
\end{aligned}
\label{eq:focalloss}
\end{equation}
\end{small}

\subsubsection{Model Train \& Testing}

Once the optimal hyperparameters are selected, the final AD model is trained utilizing both the training and the validation datasets. During this step, the model minimizes the chosen loss function to accurately differentiate between normal and anomalous patterns. After training, the model’s performance is evaluated on the test dataset using metrics suited for AD: precision, recall, and F1-score (Equations \ref{eq:prec}, \ref{eq:rec}, and \ref{eq:f1}). These metrics rely on key values: True Positives (TP), representing correctly identified anomalies; False Positives (FP), which are normal instances mistakenly identified as anomalies; True Negatives (TN), correctly identified normal instances; and False Negatives (FN), where anomalies go undetected. Precision reflects the proportion of correctly identified anomalies among all detections, while recall measures the model’s capability to identify all actual anomalies. The F1-score, as the harmonic mean of precision and recall, provides a balanced measure of sensitivity and accuracy, ensuring the model performs effectively in identifying anomalies without overestimating or underestimating anomalous events.

\begin{small}
    \begin{equation}
    \text{Precision} = \frac{\text{TP}}{\text{TP} + \text{FP}}
        \label{eq:prec}
    \end{equation}
\end{small}
\begin{small}
    \begin{equation}
    \text{Recall} = \frac{\text{TP}}{\text{TP} + \text{FN}}        \label{eq:rec}
    \end{equation}
\end{small}

\begin{small}
    \begin{equation}
    F1 = 2 \cdot \frac{\text{Precision} \cdot \text{Recall}}{\text{Precision} + \text{Recall}}
        \label{eq:f1}
    \end{equation}
\end{small}

\subsection{ShaTS xAI Module}

Having outlined the typical structure of an AD framework, the proposed contribution, ShaTS, is now introduced. ShaTS is a plug-and-play xAI module designed for use within frameworks with ML/DL models that handle time-series data, particularly in the context of AD for IIoT. ShaTS is open-source and its code is available at GitHub \footnote{\url{https://github.com/CyberDataLab/ShaTS}}.

A key innovation of ShaTS is its ability to overcome the limitations of existing xAI methods. Although ML/DL models effectively leverage the sequential nature of time-series data to boost performance, most xAI methods often overlook these temporal relationships. ShaTS bridges this gap by introducing feature grouping before computing Shapley values. ShaTS xAI module offers three distinct grouping strategies to address various explanatory requirements, providing contextually relevant and actionable insights.

The ShaTS xAI module consists of three main components: Background Dataset Creation, Grouping Strategy Selection, and ShaTS Value Computation.

\subsubsection{Background Dataset Creation}

The creation of a background dataset is a fundamental step in Shapley-based xAI methods, allowing models to handle absent features by approximating the marginal distribution of the model’s output. This dataset, constructed from a representative sample of the training data, captures typical operational patterns and feature dependencies within the model's domain.

In ShaTS, the background dataset is used to estimate the coalition value function $v(S,x_*)$. This function estimates the model’s output using features in subset $S$, while substituting values from the background dataset for features outside $S$. The coalition value is calculated by averaging the model's predictions over these combinations, as described in Equation \ref{eq:support}.  In this Equation, $K$ is the number of samples in the background dataset, $x_*^S$ denotes the values of the features in $S$ for the instance $x_*$, and $x_k^{\overline{S}}$ represents the values of the features not in $S$ from the $k$-th sample of the background dataset.

\begin{small}
\begin{equation}
v(S, x_*) \approx \frac{1}{K} \sum_{k=1}^{K} f(x_{\ast}^S, x_{k}^{\overline{S}})
\label{eq:support}
\end{equation}
\end{small}

\subsubsection{Grouping Strategy Selection}

This component is responsible for organizing features into groups before computing their ShaTS values, providing three distinct grouping strategies that align with different explainability objectives:

\begin{enumerate}
    
    \item \textbf{Temporal Grouping Strategy}: Each group contains all measurements recorded at a specific instant within the time window. This strategy is useful for identifying critical events or instants that have significantly influenced the model’s prediction.

    \item \textbf{Feature Grouping Strategy}: Each group represents the measurements of an individual feature over the time window. This strategy is particularly effective for isolating the impact of specific features on the model’s decisions.

    \item \textbf{Multi-Feature Grouping Strategy}: Each group includes the combined measurements over the time window of features that share a logical relationship or represent a cohesive functional unit. This approach is suitable for analyzing the collective impact of features that interact or are interdependent, ensuring that their combined influence is captured.

\end{enumerate}

\subsubsection{ShaTS Value Computation}

This component is responsible for calculating the ShaTS value for each group $G_i$ using either the exact or the approximate method. 

The exact method, shown in Equation \ref{eq:group_shapley_exact}, involves evaluating all subsets $T$ of groups $G$ excluding $G_i$. This approach requires evaluating $2^{|G|-1}$ coalitions for each ShaTS value, which can become computationally intensive as $|G|$ grows.

\begin{small}
\begin{equation}
\varphi_{G_i} = \sum_{T \subseteq G \setminus G_i} \frac{|T|! \, (|G| - |T| - 1)!}{|G|!} \cdot \left[ v(T \cup G_i) - v(T) \right]
\label{eq:group_shapley_exact}
\end{equation}
\end{small}

The approximation method \cite{approshap} reduces the computational effort by using a stratified sampling strategy to select a reduced subset of $m$ coalitions for each group. This calculation uses the alternative expression of the Shapley value presented in Section \ref{sec_marco}, specifically Equations \ref{eq:shapleyAlternativo1} and \ref{eq:shapleyAlternativo2}. In this way, the exact ShaTS value can be approximated as shown in Equation  \ref{eq:group_shapley_exactALT}, where each $\varphi_{G_i}^j(v)$ can be calculated following Equation \ref{eq:group_shapley_approx1}, with each $T_j^{G_i}$ representing the set of subsets of $G$ that do not contain $G_i$ and are of size $j$.

\begin{small}
\begin{equation}
\varphi_{G_i}(v) = \frac{1}{|G|} \cdot \sum_{j=0}^{|G|-1} \varphi_{G_i}^j(v)
\label{eq:group_shapley_exactALT}
\end{equation}
\end{small}

\begin{equation}
    \varphi_{G_i}^j(v) = \frac{1}{|T^j_{G_i}|} \sum_{T \in T^j_{G_i}}  \cdot \left[ v(T \cup \{G_i\}) - v(T) \right]
\label{eq:group_shapley_approx1}
\end{equation}

To approximate each ShaTS value, a reduced set \( R_{G_i}^j \subset T_{G_i}^j \) is selected for each feature group \( G_i \) and subset size \( j \). A total of \( m \) subsets is chosen to approximate \( \varphi_{G_i}(v) \), following the method described in \cite{approshap}. This total \( m \) is the sum of all \( m_j \), where each \( m_j \) indicates the number of subsets of size \( j \) considered for each feature group \( G_i \). Consequently, each reduced set \( R_{G_i}^j \) contains \( m_j \) elements, with \( m_j \) calculated according to Equation \ref{eq:mj}.

\begin{equation}
m_j = \min \{ \Biggl\lfloor \frac{m \cdot (j+1)^{\frac{2}{3}}}{\sum_{k=0}^{|G|}(k+1)^\frac{2}{3} } \Biggr\rfloor, \binom{|G|-1}{j} \}
\label{eq:mj}
\end{equation}

Once \( m_j \) is determined, the sets belonging to \( R_{G_i}^j \) are randomly selected from the larger set \( T_{G_i}^j \), which contains all the subsets of size \( j \) that exclude \( G_i \). Finally, the approximate ShaTS value for a given \( m \) is calculated according to Equation \ref{eq:group_shapley_approx}.

\begin{equation}
\begin{aligned}
\varphi_{G_i} = & \frac{1}{|G|} \sum_{j=0}^{|G|-1} \varphi_{G_i}^j(v) \approx \frac{1}{|G|} \sum_{j=0}^{|G|-1} \frac{1}{m_j} \sum_{T \in R_{G_i}^j} \left( v(T \cup \{G_i\}) - v(T) \right)
\end{aligned}
\label{eq:group_shapley_approx}
\end{equation}

\section{Experimentation}
\label{sec_experimentation}

This section includes the experimental setup and evaluation aimed at assessing both ShaTS’s ability to provide actionable explanations within an AD framework for IIoT and its overall computational efficiency. The experiments compare ShaTS against the conventional SHAP approach \cite{kernelshap}, verifying its ability to preserve temporal dependencies, provide explainable insights, and minimize resource consumption. The procedure is organized into five main parts: an overview of the SWaT dataset, data preprocessing, AD model generation, integration and evaluation of the ShaTS xAI module, and a comprehensive discussion that includes both xAI findings and resource-usage results.

\subsection{SWAT Dataset}

The Secure Water Treatment (SWaT)\cite{swat} testbed is a fully operational, scaled-down water treatment system used to benchmark cybersecurity in IIoT. Comprising six interconnected processes that replicate large-scale treatment operations, SWaT logs sensor and actuator data every second, amassing 946\,722 samples over 11 days. It includes 51 features representing physical properties, with data collection split between a seven-day normal operation baseline and four days involving 36 cyberattacks with actual physical impact.

\begin{figure}[htbp]
    \centering
    \includegraphics[width=0.8\linewidth]{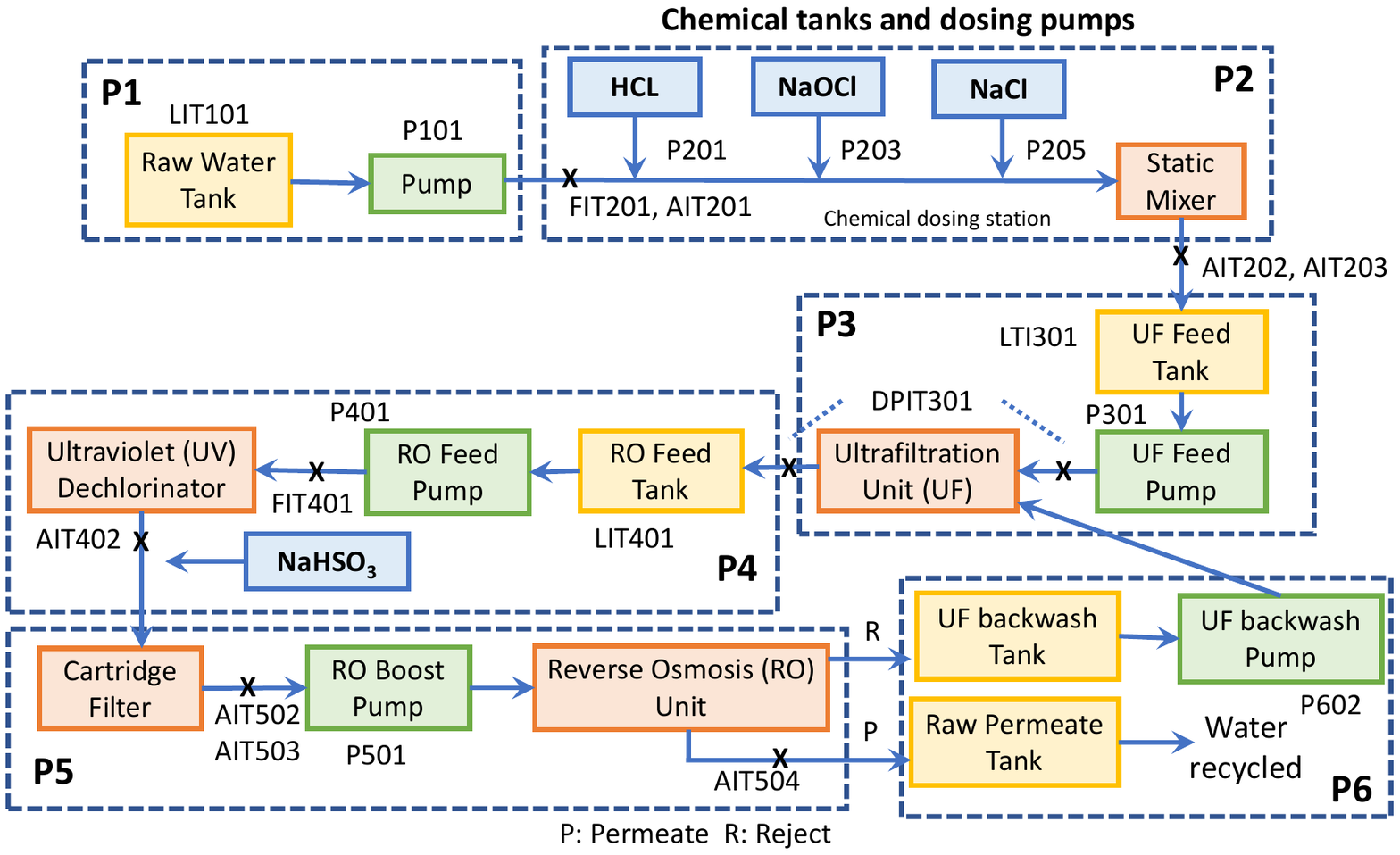}
    \caption{SWaT testbed processes overview \cite{perales2020madics}}
    \label{fig:swat_esquema}
\end{figure}

Each SWaT process models a distinct treatment phase: Raw Water Storage (P1) stores untreated water; Pre-treatment (P2) conducts quality checks and chemical dosing; Ultra Filtration (P3) removes particulates; Dechlorination (P4) neutralizes chlorine; Reverse Osmosis (P5) eliminates inorganic impurities; and Water Distribution (P6) prepares purified water for storage or distribution. Figure \ref{fig:swat_esquema} illustrates these stages, while Table \ref{table:procesos_swat} details associated sensors and actuators.

\begin{table}[htbp]
\footnotesize
\centering
\begin{tabular}{|c|c|c|}
\hline
\textbf{Process} & \textbf{Description} & \textbf{Sensors and actuators} \\
\hline
P1 & Raw Water Storage & \makecell{FIT101, LIT101, MV101 \\ P101, P102} \\
\hline
P2 & Water Pre-treatment & \makecell{AIT201, AIT202, AIT203 \\ FIT201, MV201, P201 \\ P203, P204, P205 \\ P206} \\
\hline
P3 & Ultra Filtration & \makecell{DPIT301, FIT301, LIT301 \\ MV301, MV302, MV303 \\ MV304, P302} \\
\hline
P4 & Decholination & \makecell{AIT401, AIT402, FIT401 \\ LIT401, P402, P403 \\ UV401} \\
\hline
P5 & Reverse Osmosis & \makecell{AIT501, AIT502, AIT503 \\ AIT504, FIT501, FIT502 \\ FIT503, FIT504, P501 \\ PIT501, PIT502, PIT503} \\
\hline
P6 & Water Distribution & \makecell{FIT601, P602} \\
\hline
\end{tabular}
\caption{Mapping of SWaT Processes to Associated Sensors and Actuators. Each row corresponds to a distinct stage of the water treatment system, with sensor and actuator identifiers listed for the respective process.}
\label{table:procesos_swat}
\end{table}

\subsection{Data Preprocessing}

For this experimentation, the labeled subset of the SWaT dataset was utilized, containing annotations for both normal and attack instances. This stage focused on preparing two configurations, each one adapted to the requirements of the experimentation.

\subsubsection{Data Cleaning}
During data inspection, no noise or missing values were identified in the SWaT dataset, making additional cleaning steps, such as filtering or imputation, unnecessary. This finding made it possible to use the raw data in its original form without requiring further adjustments.

\subsubsection{Dataset Creation}

The SWaT dataset was initially divided into segments of 1\,000 time steps. Within each segment, 80\% of the data was allocated for training and validation, with an internal split of 80\% for training and 20\% for validation, while the remaining 20\% was reserved for testing. To prevent data leakage, a padding of 50 time steps was added between each of these subsets. The test dataset was designed to include a representative portion of impactful attacks, with 28 out of the 36 attacks containing segments within it.

From the initial dataset split, two configurations were created, each tailored to different experimental objectives. The first configuration was designed for experiments involving temporal grouping, applying a stride of one between windows. This approach enables finer-grained analysis of the start and end of attacks, as each window advances by a single time step, allowing detailed observation of how attacks gradually emerge or diminish across windows. This dataset was exclusively used for temporal ShaTS experiments and not for model training. The second configuration was prepared for ShaTS experiments involving sensor/actuator and process grouping, using a stride of three between windows. This configuration was also employed for model training. In both setups, each window spans 10 instants and has assigned the same label as the final instant within the window.

\subsubsection{Feature Preparation}

To optimize the dataset for model training, a variance analysis was first conducted to identify and remove inactive sensors and actuators. Originally, the dataset comprised 51 sensors and actuators; however, seven (P202, P301, P401, P404, P502, P601, and P603) exhibited zero variance and were excluded. This filtering resulted in 44 active sensors and actuators. Among these, 25 produced continuous data, while the remaining 19 yielded categorical data. The 25 continuous features were normalized, standardizing them to have a mean of zero and a standard deviation of one to ensure consistency and to prevent any single feature from disproportionately influencing the model. Meanwhile, the 19 categorical features were transformed via one-hot encoding, generating 44 binary features. In total, this process produced a dataset with 69 features: 25 continuous features corresponding to the continuous sensors and 44 binary features representing the categorical sensors.

\subsection{Model Generation}

To ensure accurate AD in the SWaT dataset, a Stacked Bidirectional LSTM model was implemented, fine-tuned, and evaluated for its reliability in identifying attacks.

\subsubsection{AD Model Selection}

A Stacked Bidirectional LSTM architecture was selected to capture complex temporal dependencies in the data. This model leverages LSTM units to identify both short- and long-term patterns while bidirectional processing enhances its ability to analyze temporal relationships in forward and backward directions. The architecture includes multiple LSTM layers stacked to learn hierarchical representations, followed by a fully connected output layer with a sigmoid activation function that maps predictions to a probability score between 0 and 1. A classification threshold of 0.5 is applied, where scores above this value indicate anomalies, and scores below classify the window as normal. Although various AD architectures could be explored, this study employs a Stacked Bidirectional LSTM as a reliable baseline for evaluating the proposed ShaTS xAI module.

\subsubsection{Model Hyperparameter Tuning \& Training Setup}

While optimizing AD performance is not the main focus of this work, the model must perform with sufficient accuracy to ensure that subsequent xAI analysis provides reliable insights. To this end, hyperparameter tuning was conducted using a Grid Search strategy, systematically testing each combination of values within the defined space to identify the best configuration. The optimized hyperparameters included the number of hidden dimensions, the LSTM layers, the dropout rate, the learning rate, and the training epochs, as outlined in Table \ref{tab:gridsearch}. Additionally, the key hyperparameters of Focal Loss ($\alpha$ and $\gamma$) were included. Focal Loss, particularly suited to address the dataset’s significant class imbalance, down-weights well-classified instances and emphasizes difficult-to-classify ones, mitigating the disproportionate influence of normal instances and enhancing the model's sensitivity to shorter attacks that might otherwise be overlooked.

\begin{table}[ht] 
\centering 
\footnotesize
\begin{tabular}{|c|c|} 
\hline 
\textbf{Hyperparameter} & \textbf{Values Tested} \\ \hline 
Hidden Dimensions & 16, \textbf{32}, 64, 128 \\ \hline
Number of Layers & \textbf{2}, 3, 5 \\ \hline 
Dropout & 0, \textbf{0.2}, 0.5 \\ \hline
Learning Rate & \textbf{0.001}, 0.0005 \\ \hline
Number of Epochs & \textbf{5} \\ 
\hline 
$\alpha$ (Focal Loss)& 0.5, \textbf{1} \\
\hline
$\gamma$ (Focal Loss)& 1, \textbf{2} \\
\hline
\end{tabular} 
\caption{Hyperparameter Grid Search for selected AD model. The values in bold represent the best found values.} 
\label{tab:gridsearch} 
\end{table}

\subsubsection{Model Train \& Testing}

After fine-tuning, the final configuration of the model utilized the optimal hyperparameters from the grid search, as highlighted in Table \ref{tab:gridsearch}. With these values, the model achieved a test loss of 0.0024, an accuracy of 98.13\%, a precision of 98.17\%, a recall of 93.09\%, and an F1-score of 95.44\%. These results indicate that the anomaly detector is well-calibrated, with strong precision ensuring a low rate of false positives and high recall enabling effective identification of genuine anomalies. The combination of these metrics, as reflected in the F1-score, demonstrates the model's balanced performance in handling both aspects of classification.

The model was also evaluated based on its ability to detect individual attacks. Out of the 28 attacks included in the test dataset, the model successfully detected 21, with 19 of these detected with perfect accuracy, which means that the model correctly identified all time windows within those attacks. These metrics demonstrate the reliability of the model in detecting attacks, confirming that it provides a solid basis for further explanability analysis using the ShaTS xAI module.

\subsection{ShaTS xAI Module}

The ShaTS xAI Module provides critical insights into why an anomaly has been detected within an existing AD framework. The module operates through three key steps: Background Dataset Creation, Grouping Strategy Selection, and ShaTS Value Computation.

\subsubsection{Background Dataset Creation} 

A background dataset of 500 instances was created as a stratified subset of the training data, ensuring that the proportion of normal and anomalous traffic remains representative of the broader dataset. In particular, 12.1\% of these instances are anomalous, mirroring the ratio observed in the original data. This dataset is used to approximate the value of each coalition in ShaTS computations, thereby reducing bias and capturing the typical behavior of the system. By preserving sufficient normal and anomalous samples, the background dataset enables more accurate estimation of the model’s output when features are excluded or substituted, ultimately leading to more reliable ShaTS values.

\subsubsection{Grouping Strategy Selection}

To meet the needs of the subsequent experiments, three distinct groupings of the AD model's features were constructed. Each grouping corresponds to one of the ShaTS strategies defined in Section \ref{sec_proposed}. 
\begin{itemize}
    \item \textbf{Temporal Grouping}: This configuration uses the Temporal Grouping Strategy, where each group corresponds to the measurements recorded at a specific instant within the time window. This approach allows for the identification of critical moments that significantly influence the model’s predictions, thereby directly pinpointing the onset or termination of anomalous behavior. As each time window contains 10 instants, the resulting number of groups is $|G| = 10$.
 
    \item \textbf{Sensor/Actuator Grouping}: For this configuration, the Multi-Feature Grouping Strategy is employed to group all measurements related to individual sensors and actuators across the time window. Each group aggregates the contributions of all features associated with a single sensor or actuator. Notably, some sensors or actuators in the SWaT dataset are represented by multiple features due to one-hot encoding. This grouping is crucial for identifying which specific sensor or actuator is driving an anomaly, thereby offering actionable insights for targeted interventions. After feature preparation, the SWaT dataset comprises 44 active sensors and actuators, resulting in $|G| = 44$.

    \item \textbf{Process Grouping}: The Multi-Feature Grouping Strategy is also applied in this configuration to aggregate measurements from all sensors and actuators associated with the same industrial process. By evaluating the combined influence of these features, it provides a higher-level view of operational performance and localizes anomalies at the process level, which is especially valuable when individual sensor signals are ambiguous. The SWaT dataset defines six processes, so $|G| = 6$. The mapping of sensors and actuators to these processes is detailed in Table \ref{table:procesos_swat}.

\end{itemize}

\subsubsection{ShaTS Value Computation}

For all three levels of grouping, the approximate computation method is selected. This method requires defining a parameter \( m \), representing the number of coalition subsets to be used for estimating each ShaTS value. To ensure \( m \) is dependent on the number of groups being evaluated, $m$ is set as \( m = 20 \cdot |G| \). Thus, for temporal grouping, \( m = 200 \); for sensor/actuator grouping, \( m = 880 \); and for process grouping, \( m = 120 \). If the exact computation method had been chosen, \( m \) would instead equal \( 2^{|G|} - 1 \), which would be significantly more computationally expensive and less suitable for real-time industrial requirements.

\subsection{Experimental Results}

The experimental procedure was divided into two parts. The first set focused on the reliability of ShaTS’s explanations, while the second set investigated resource usage and execution time. In both cases, ShaTS was compared against the conventional method of computing Shapley values via KernelSHAP \cite{kernelshap} and aggregating them post hoc. All experiments were carried out on a workstation equipped with an Intel\textsuperscript{\textregistered} Core\textsuperscript{TM} i7-14700K processor (28 cores) and an NVIDIA GeForce RTX 4080 GPU with 16 GB of VRAM. The environment included CUDA 12.0.140 and SHAP library version 0.46.0.

\subsubsection{xAI Experimentation}

ShaTS’s ability to deliver accurate and actionable explanations was evaluated through two experiments. The first one determined whether ShaTS could identify anomalous instants in time windows containing both normal and anomalous observations, while the second assessed its capacity to pinpoint the specific physical sensor/actuator or process under attack. In each case, ShaTS was directly compared against KernelSHAP with post hoc feature grouping. To ensure fairness, KernelSHAP was configured to use the same background dataset as ShaTS. Regarding the number of subsets used for KernelSHAP, the library's auto option was selected for these experiments, yielding $m = 2 \times \text{features} \times \text{windowSize} + 2048 = 3428$.

To support the experimental analysis, custom visualizations were developed and applied to both ShaTS and post hoc SHAP approaches. In these graphs, the x-axis represents consecutive time windows, while the left y-axis shows the grouped features, whether by time instants, sensor/actuator, or process, and the right y-axis displays the model's predicted anomaly probability. Each point at $(i,j)$ indicates the importance of feature group $j$ for window $i$, with color intensity reflecting the importance value. Red hues indicates positive contributions (deeper reds for stronger influence), while blue hues denotes negative contributions (darker blues for stronger non-anomalous influence). Uncolored areas represents groups with negligible influence. The purple line depicts the predicted anomaly probability, with a horizontal threshold at 0.5 to classify windows as anomalies (above the threshold) or normal (below).

\textbf{Experiment 1: Temporal Consistency Analysis}

This experiment, conducted on the test dataset with a stride of one, evaluated ShaTS’s ability to highlight specific instants within windows that indicated the start or end of an attack. Since each window was labeled according to its final instant, it could have contained both anomalous and normal instants as the attack begins or ends.

To evaluate the effectiveness of ShaTS in this experiment, the Temporal Grouping Strategy was applied so that each group corresponded to all the measurements recorded at a specific instant within the window. This approach computed ShaTS value for each time instant as a group, preserving the sequential structure of the data. In contrast, the conventional method calculated SHAP values by evaluating each feature independently at each time instant and then aggregated them post hoc to derive the overall contribution of that instant.

\begin{figure}[h]
    \centering
    \begin{minipage}{0.49\linewidth}
        \centering
        \includegraphics[width=\linewidth]{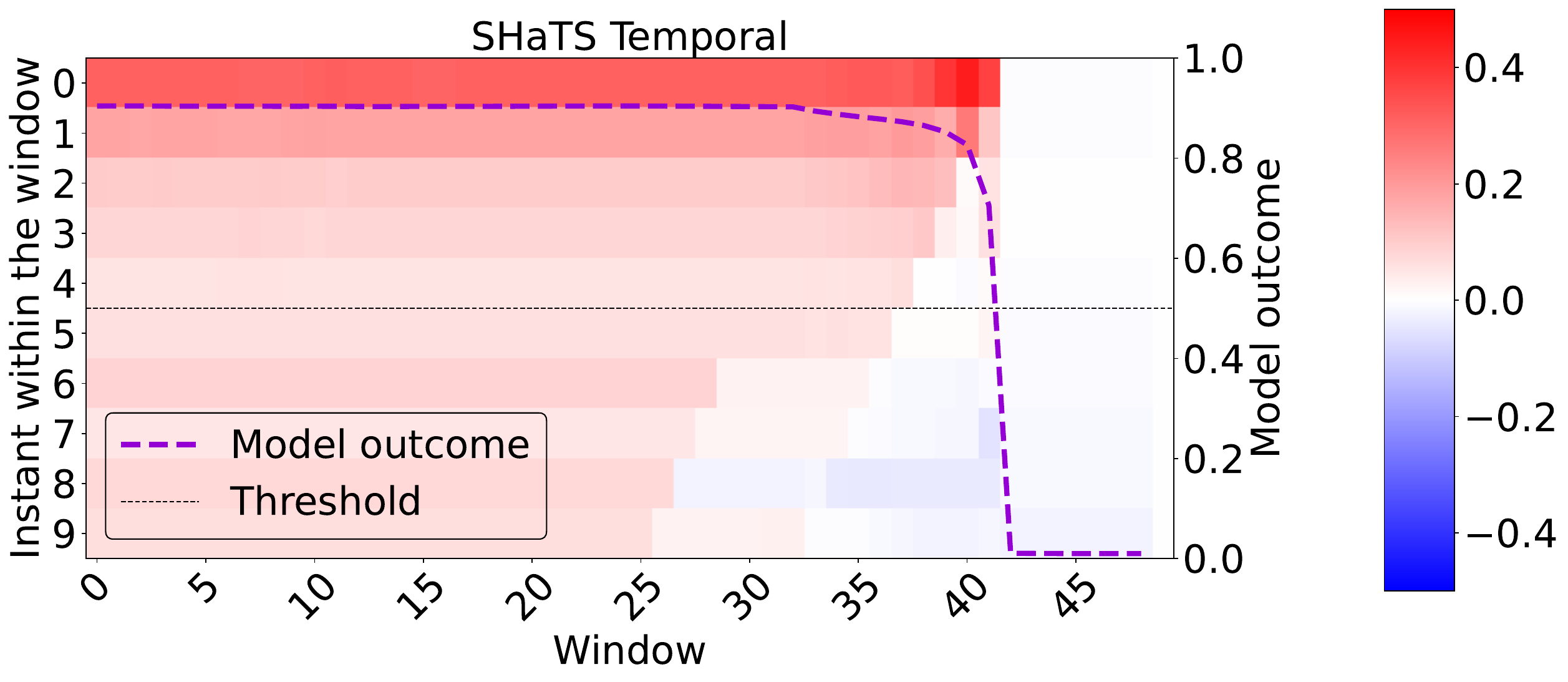}
        \label{fig:at1_time_left}
    \end{minipage}
    \hfill
    \begin{minipage}{0.49\linewidth}
        \centering
        \includegraphics[width=\linewidth]{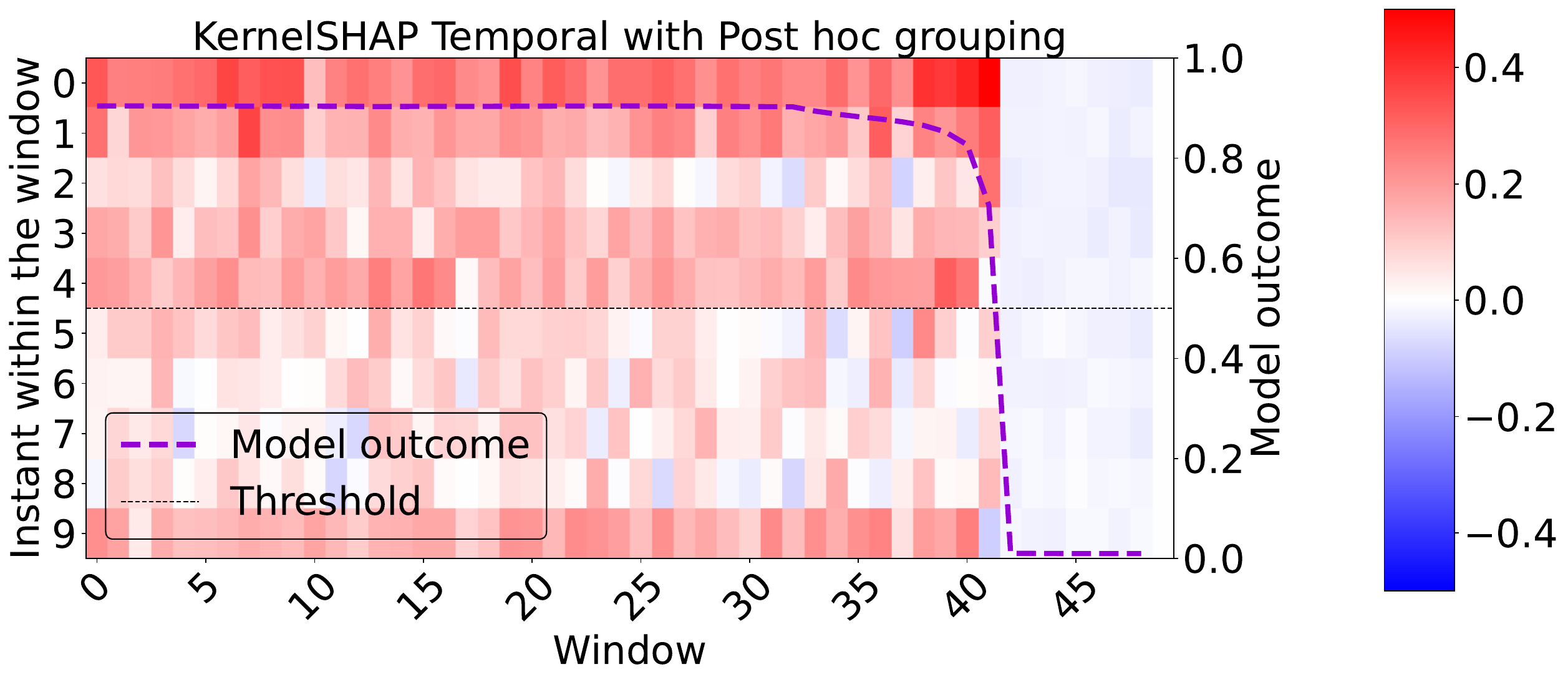}
        \label{fig:at1_time_rigth}
    \end{minipage}
\caption{Comparison of ShaTS (left) and post hoc SHAP (right) for the end of Attack 1 at the temporal level.}
\label{fig:at1_time}
\end{figure}

Due to the Data Preprocessing, only 9 of the 21 correctly classified attacks had either their onset or termination captured within the test dataset. In these cases, ShaTS’s temporal grouping effectively isolated the critical instants: at the start of an attack, the later instants in each window (the anomalous ones) contributed most to the model's predictions, while at the end of an attack, the earlier instants (the non-anomalous ones) began to dominate. This shifting pattern created a diagonal effect in the heatmaps, which was more pronounced due to the stride of one, allowing gradual observation of transitions from normal to anomalous instants (or vice versa) as the attack unfolded.
Figure \ref{fig:at1_time} compares ShaTS with Temporal Grouping Strategy (left) with conventional post hoc KernelSHAP (right) for the last samples of Attack 1. On the left, ShaTS revealed a clear shift in influence as the anomalous instants diminished and the model’s predicted anomaly probability (indicated by the purple line) dropped below the threshold. Notably, the model continued to predict an anomaly as long as at least one anomalous instant was present. In the last window classified as anomalous, ShaTS accurately predicted that only the earliest instant (instant 0) contributed significantly, indicating that the attack was coming to an end. In contrast, the post hoc SHAP visualization on the right did not capture this transition as clearly, making it more difficult to pinpoint exactly which instants within the window drove the predictions.

\textbf{Experiment 2: Attack Source Identification}

This second experiment evaluated ShaTS’s ability to pinpoint the specific physical element (sensor/actuator) or industrial process responsible for an attack. Such granularity was crucial in IIoT environments, as it enabled operators to target the affected component directly and implement more precise mitigation measures.

To achieve this, two grouping strategies were employed in ShaTS. First, the Sensor/Actuator Grouping aggregated all features corresponding to a single sensor or actuator across the time window, allowing ShaTS to isolate which device was driving the anomaly. Second, the Process Grouping aggregated all sensors and actuators associated with a given industrial process (Table \ref{table:procesos_swat}), providing a higher-level view of which stage in the water treatment cycle was compromised. By contrast, conventional SHAP computed feature attributions for each input independently and then applied a post hoc aggregation step to approximate sensor-level or process-level importance.

For each attack, the ShaTS xAI module computed importance values for all groups over every time window. The relative contribution of a group was determined by normalizing its ShaTS value by the sum of the values of all groups in that window. These normalized contributions were then averaged over the entire attack, yielding an overall importance score for each sensor/actuator (or process). The top-ranked component, defined as the one with the highest average contribution, was considered the primary source of the anomaly. The same procedure was applied to conventional SHAP with post hoc grouping to ensure a fair comparison.

Table \ref{table:exp} summarizes the results for each detected attack, showing the actual attacked component alongside the top-ranked sensor/actuator and process as identified by both ShaTS and post hoc SHAP. Of the 21 attacks correctly detected by the AD model, ShaTS correctly identified the affected sensor or actuator in 14 cases, whereas the conventional method succeeded in only 6. At the process level, ShaTS located the correct process in 19 cases compared to only 8 identified by post hoc SHAP.

\begin{figure}[h]
    \centering
    \begin{minipage}{0.49\linewidth}
        \centering
        \includegraphics[width=\linewidth]{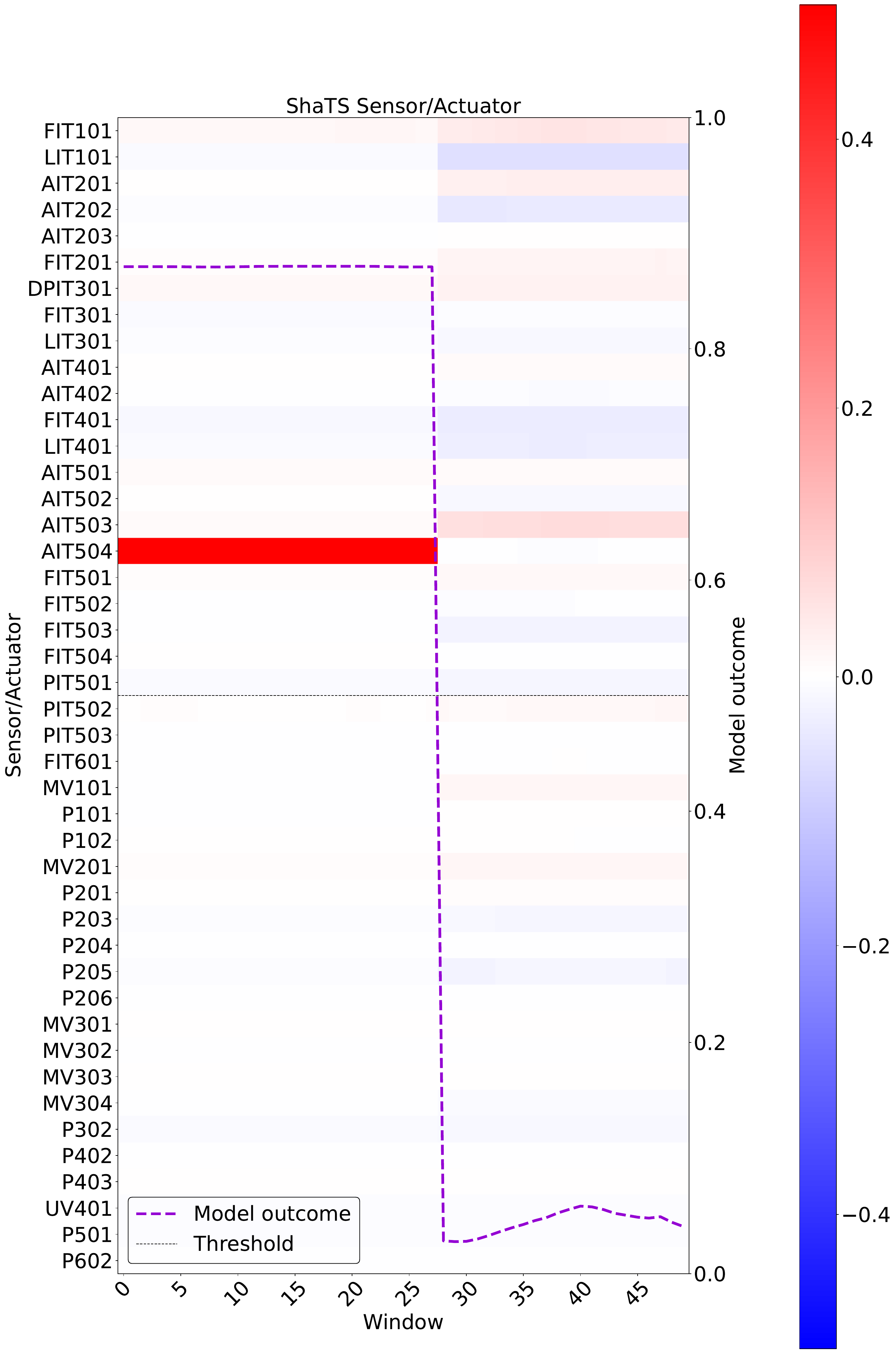}
        \label{fig:ataque19feature_left}
    \end{minipage}
    \hfill
    \begin{minipage}{0.49\linewidth}
        \centering
        \includegraphics[width=\linewidth]{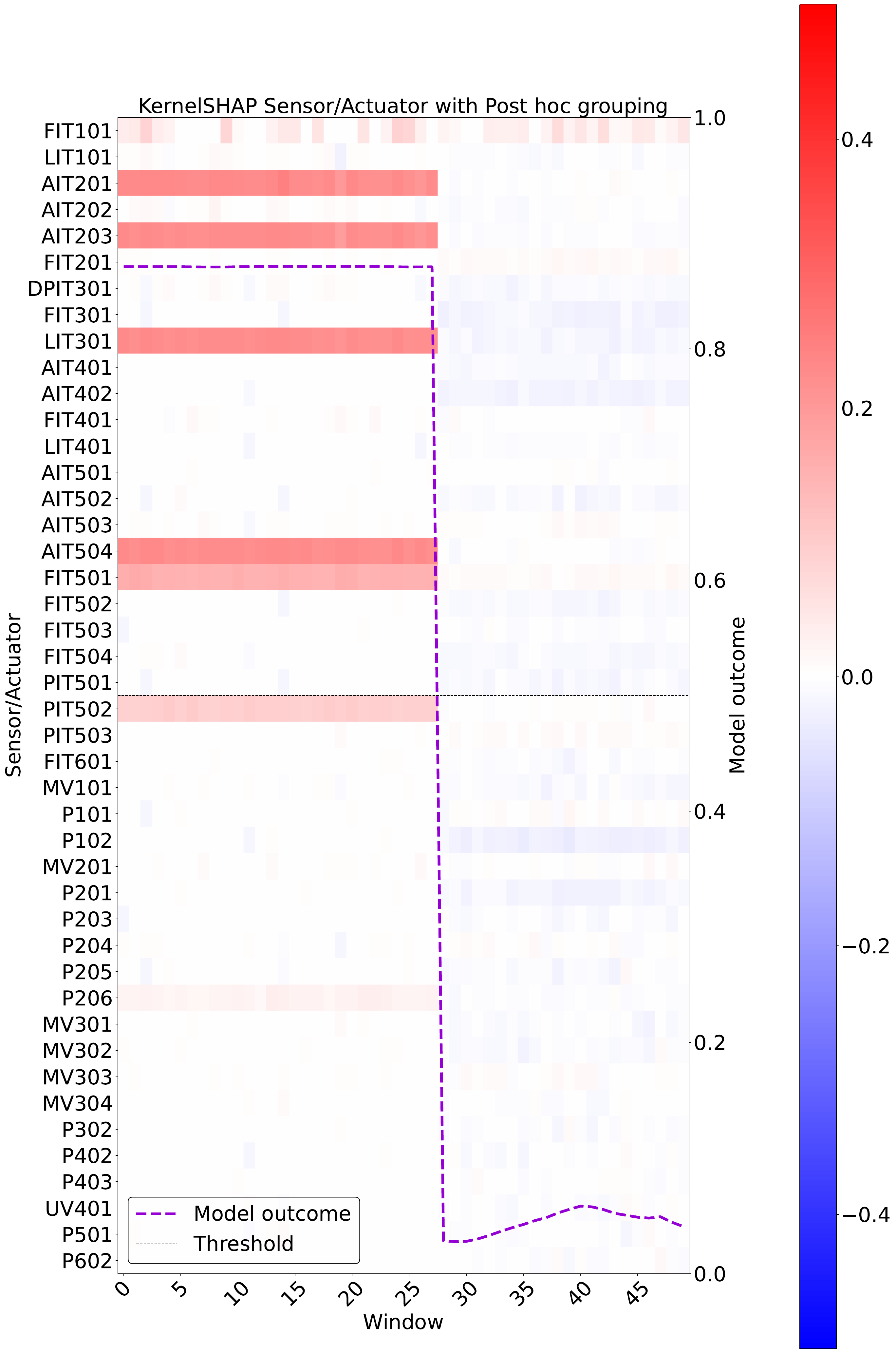}
        \label{fig:ataque19feature_rigth}
    \end{minipage}
    \caption{Comparison of ShaTS (left) and post hoc SHAP (right) for the end of Attack 19 at the sensor/actuator level.}
    \label{fig:ataque19feature}
\end{figure}

\begin{figure}[h]
    \centering
    \begin{minipage}{0.49\linewidth}
        \centering
        \includegraphics[width=\linewidth]{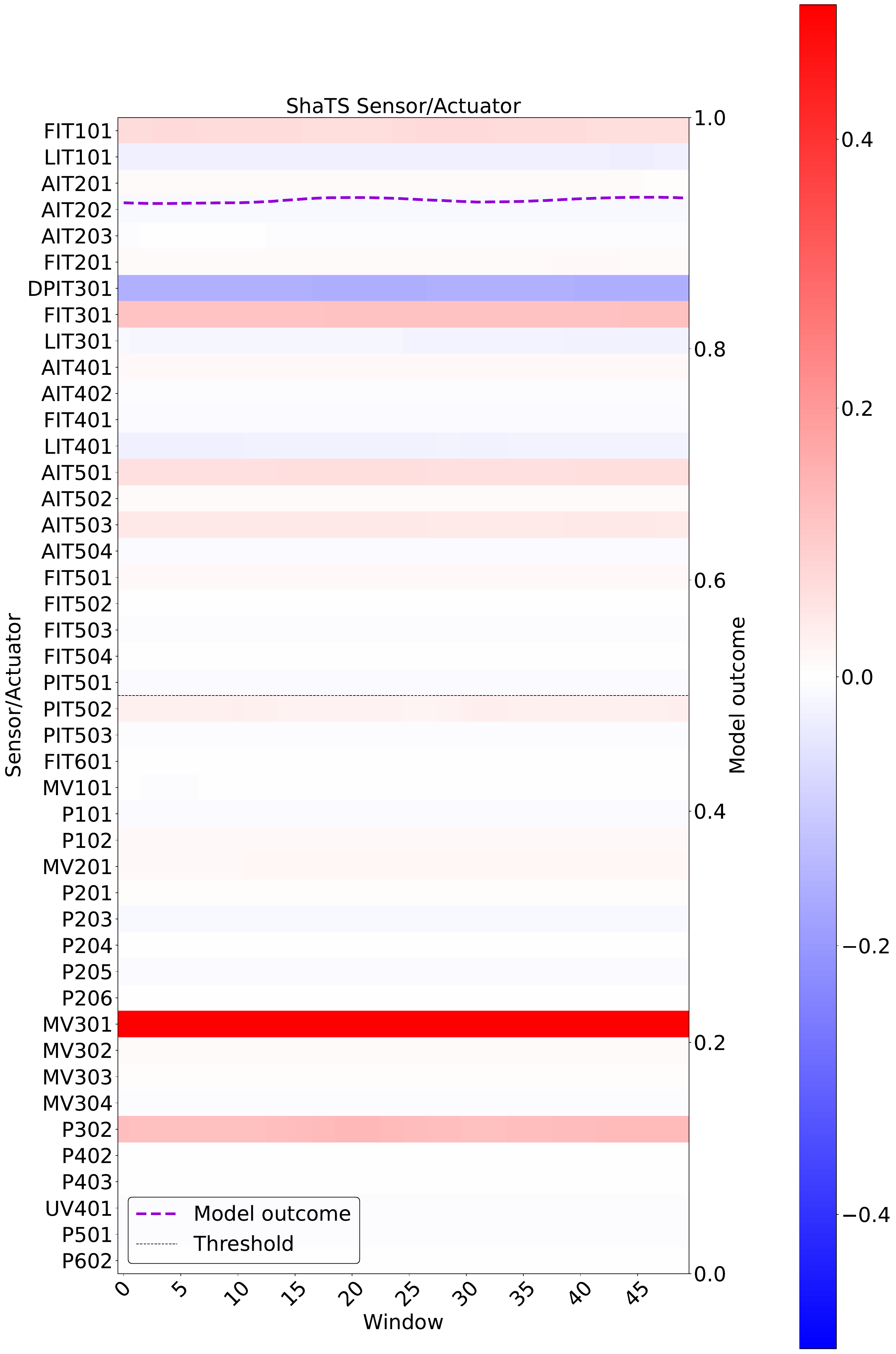}
        \label{fig:ataque17feature_left}
    \end{minipage}
    \hfill
    \begin{minipage}{0.49\linewidth}
        \centering
        \includegraphics[width=\linewidth]{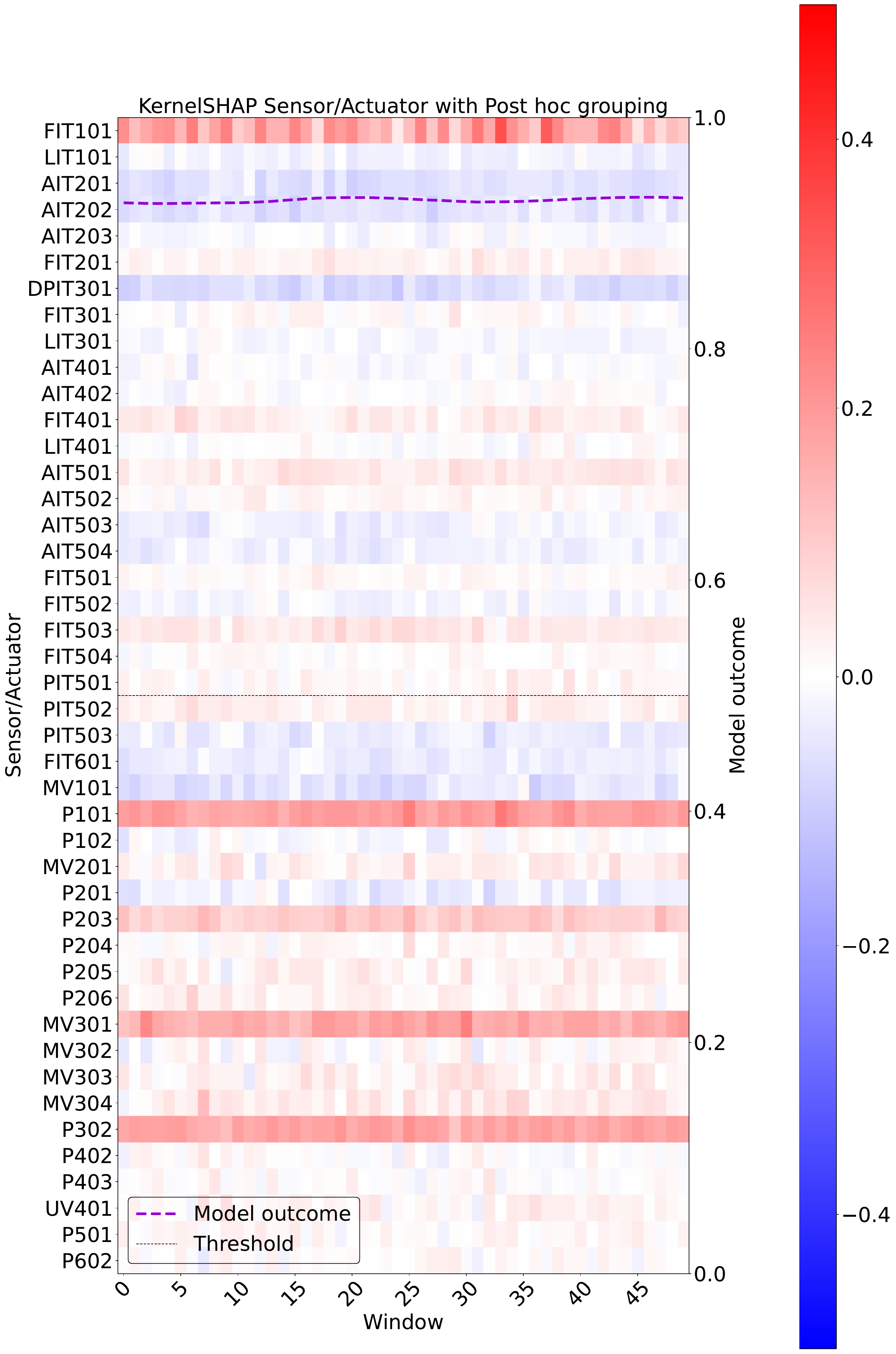}
        \label{fig:ataque17feature_rigths}
    \end{minipage}

    \caption{Comparison of ShaTS (left) and post hoc SHAP (right) for Attack 17 at the sensor/actuator level.}

    \label{fig:ataque17feature}
\end{figure}

\begin{figure}[h]
    \centering
    \begin{minipage}{0.49\linewidth}
        \centering
        \includegraphics[width=\linewidth]{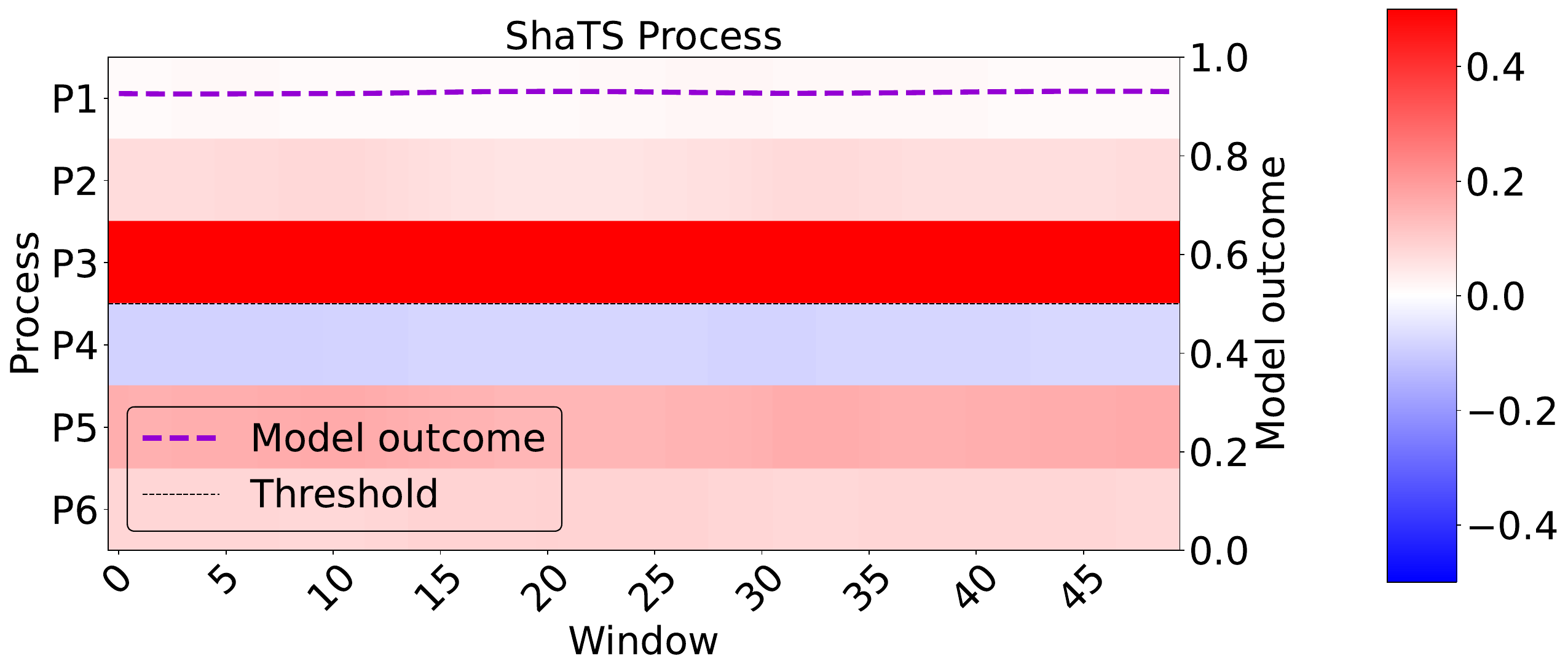}
        \label{fig:ataque17process_left}
    \end{minipage}
    \hfill
    \begin{minipage}{0.49\linewidth}
        \centering
        \includegraphics[width=\linewidth]{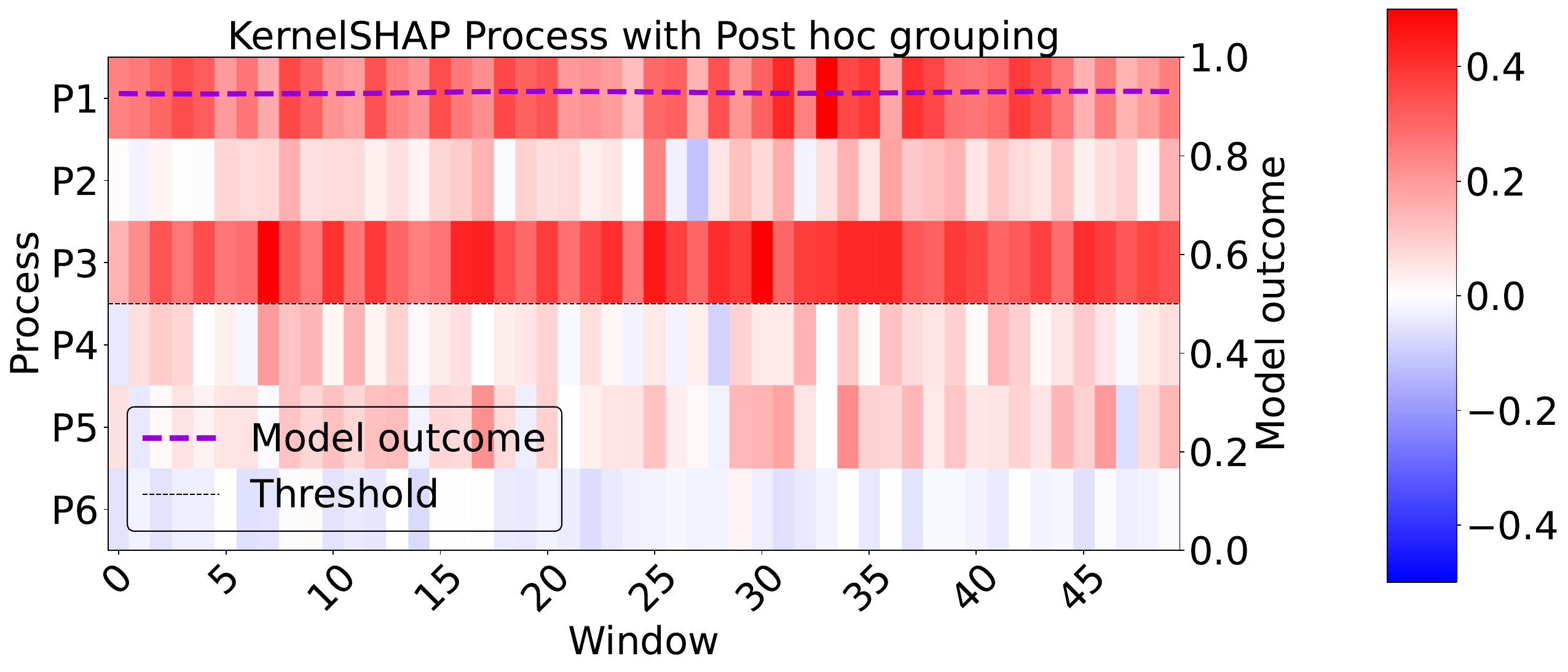}
        \label{fig:ataque17process_rigth}
    \end{minipage}
    \caption{Comparison of ShaTS (left) and post hoc SHAP (right) for Attack 17 at the process level.}

    \label{fig:ataque17process}
\end{figure}

Several figures illustrates these findings in detail. Figure~\ref{fig:ataque19feature} compares ShaTS (left) and post hoc SHAP (right) at the sensor/actuator level for the last windows of Attack 19, where the true attacked sensor was AIT504. ShaTS assigned an average importance of 91.7\% to AIT504, creating a clear red band in the heatmap and leaving no ambiguity about the anomaly’s explanation. In contrast, post hoc SHAP distributed moderate importance across many sensors, with AIT504 reaching only 17.8\%, making it more difficult to isolate the primary cause. A similar contrast appeared in the middle of Attack 17. Figure~\ref{fig:ataque17feature} presents the sensor-level attributions, whereas Figure~\ref{fig:ataque17process} shows those at the process level. Although neither method precisely pinpointed MV303 (the true target) at the sensor level, ShaTS attributed the anomaly mostly to MV301, whereas SHAP scattered responsibility among multiple features without clearly highlighting a single dominant sensor. Nonetheless, the process-level grouping in ShaTS still managed to identify Process P3 with a 70.8\% importance score, whereas post hoc SHAP gave P3 only 39.4\%.

Overall, these observations demonstrated that ShaTS’s a priori grouping strategy yielded more focused and actionable explanations than post hoc SHAP. By incorporating temporal dependencies directly into the Shapley value computation, ShaTS avoided the fragmentation of attributions that arose from conventional methods, enabling IIoT operators to respond swiftly and effectively to emerging threats.

\subsubsection{Resources experimentation}

This subsection presents a detailed comparison of KernelSHAP and ShaTS from the standpoint of computational resource usage and efficiency. Two separate experiments were conducted: the first measured VRAM/RAM consumption and execution time as functions of the number of coalition subsets and the size of the background dataset, while the second examined execution time for various AD models trained with different window sizes. These experiments provided a comprehensive view of each method’s scalability and real-time feasibility in IIoT environments.

\textbf{Experiment 3: Number of subsets and background size impact}

This experiment assessed the impact of the number of coalition subsets and background dataset size on resource consumption and execution time for both ShaTS (with sensor/actuator grouping) and conventional SHAP (with post hoc sensor/actuator grouping). The evaluation was conducted using the previously trained AD model that processed inputs of size $10 \times 69$, where each 10-step time window contained the 69 features that represented the measurements of 44 sensors and actuators over a 10-second period. Given the SWaT dataset’s one-second sampling rate, any xAI method that requiring more than one second per window would be unsuitable for real-time deployment.

\begin{figure}[htbp]
    \centering
    \begin{minipage}{0.49\linewidth}
        \centering
        \includegraphics[width=\linewidth]{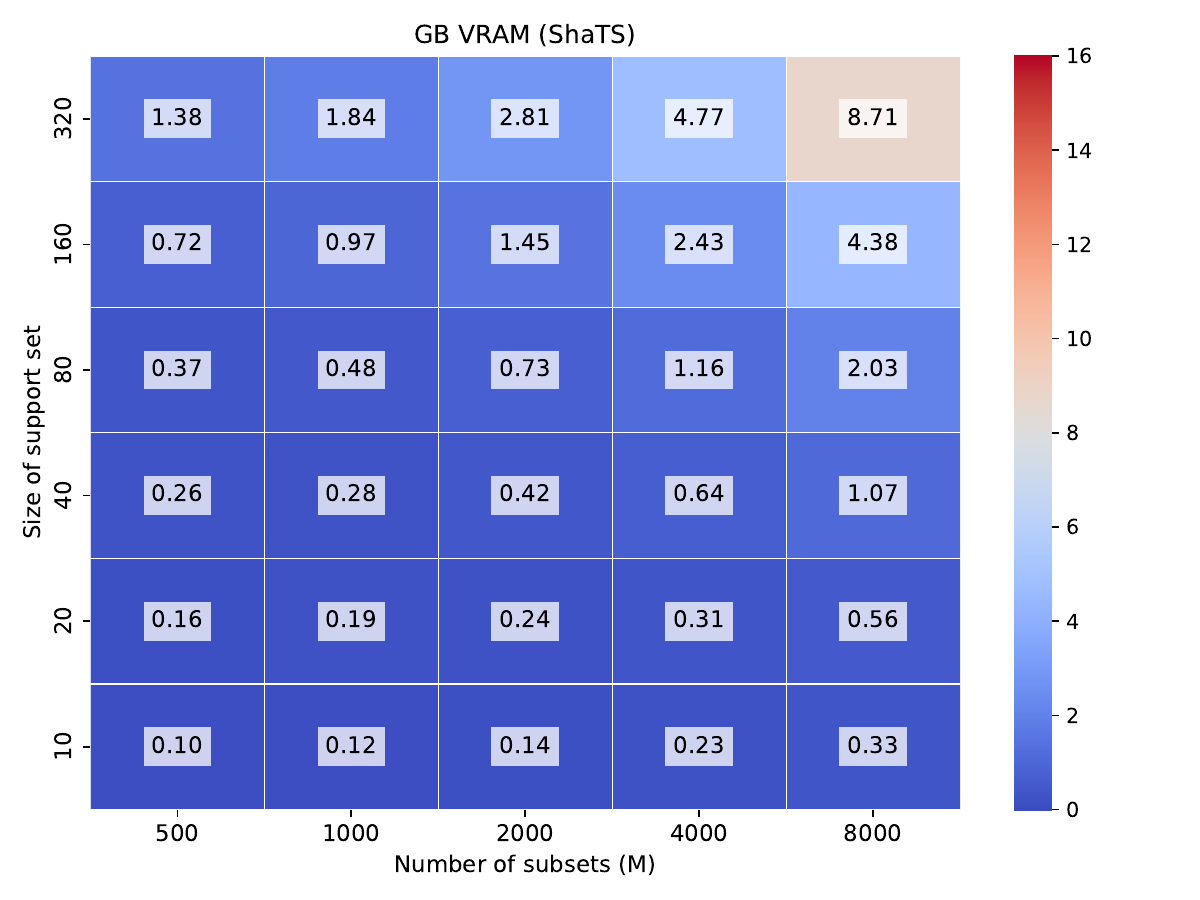}
        \label{fig:vramUsage_left}
    \end{minipage}
    \hfill
    \begin{minipage}{0.49\linewidth}
        \centering
        \includegraphics[width=\linewidth]{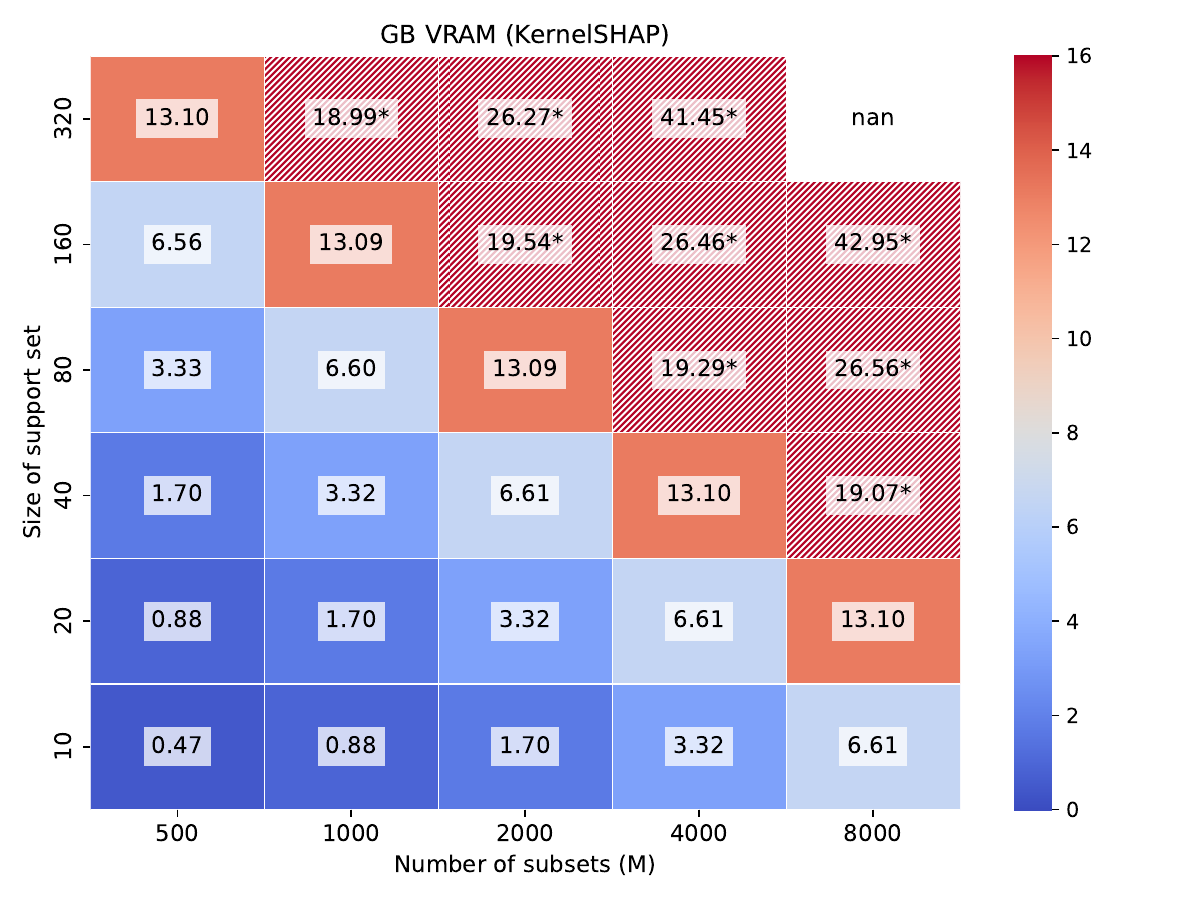}
        \label{fig:vramUsage_right}
    \end{minipage}

\caption{Comparison of VRAM consumption for ShaTS (left) and post hoc SHAP (right) when processing 100 windows, varying both the number of coalition subsets ($M$) and the background dataset size. Dashed entries indicate configurations that exceeded the 16GB GPU limit, requiring AD model and data to be offloaded to the CPU.}
    
    \label{fig:vramUsage}
\end{figure}

The number of coalition subsets was tested at five levels: 500, 1\,000, 2\,000, 4\,000 and 8\,000, while the background dataset size was varied across six configurations: 10, 20, 40, 80, 160, and 320. Although KernelSHAP did not perform Shapley computations on the GPU, it could utilize AD models that were both allocated and evaluated in GPU. However, as shown in Figure~\ref{fig:vramUsage}, SHAP’s VRAM consumption increased significantly with larger configurations, frequently exceeding the 16 GB GPU limit. In such cases, computations fell back to system RAM, as indicated by dashed and asterisk-marked entries. In the most demanding configuration, with 8\,000 coalition subsets and a background dataset size of 320, KernelSHAP exceeded the available RAM, making execution infeasible. In contrast, ShaTS consistently operated within GPU memory limits, preventing the need for CPU-based execution.

\begin{figure}[h]
    \centering
    \begin{minipage}{0.49\linewidth}
        \centering        \includegraphics[width=\linewidth]{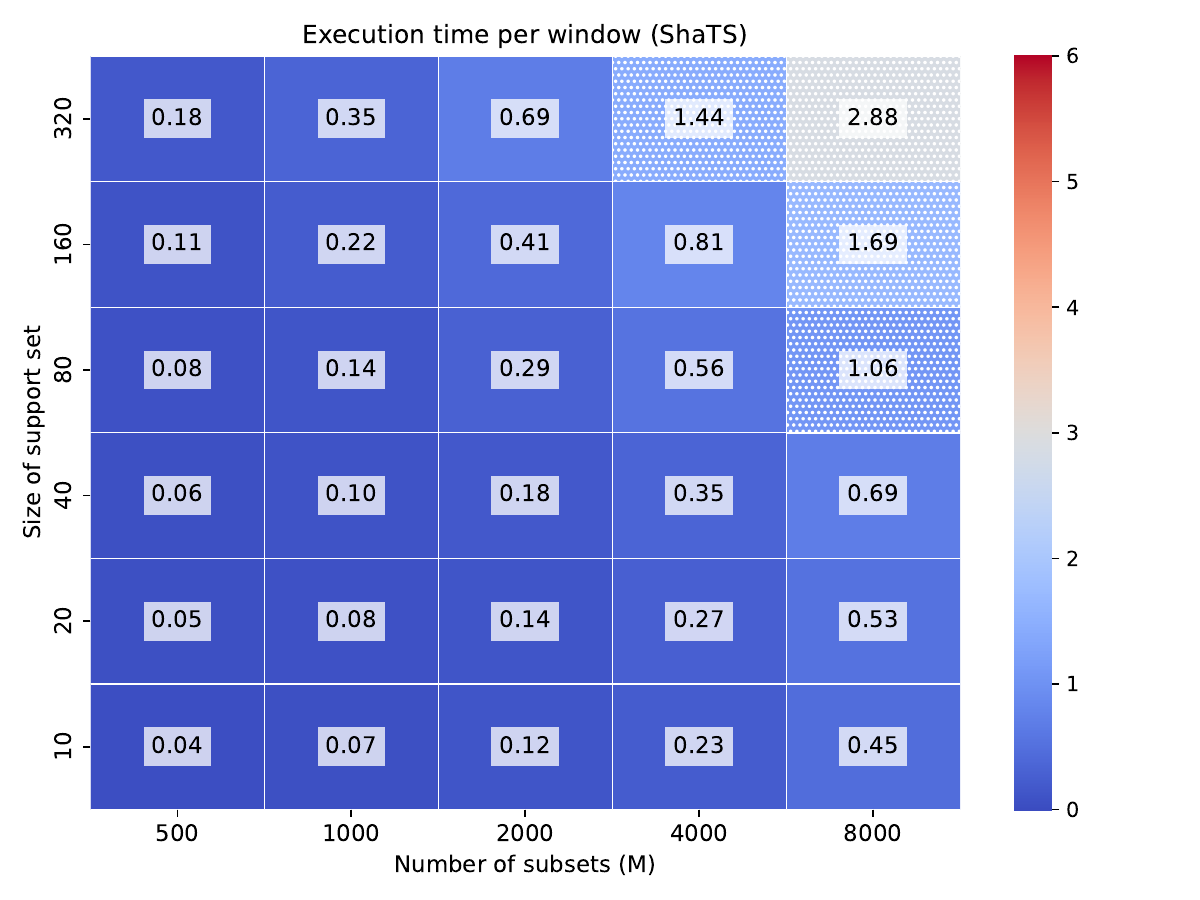}
        \label{fig:153_2}
    \end{minipage}
    \hfill
    \begin{minipage}{0.49\linewidth}
        \centering
        \includegraphics[width=\linewidth]{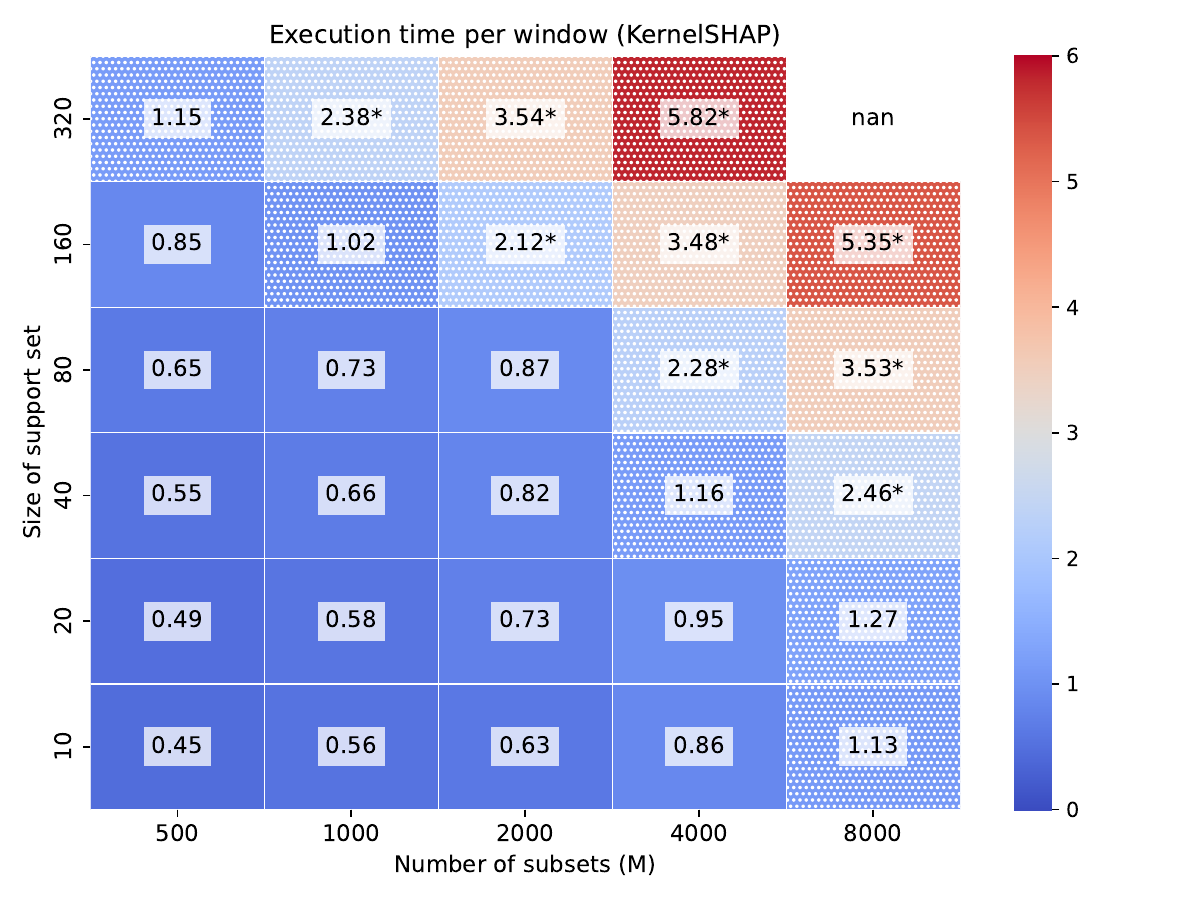}
        \label{fig:153_1}
    \end{minipage}

\caption{Comparison of mean execution time per window for ShaTS (left) and post hoc SHAP (right) when processing 100 windows, varying both the number of coalition subsets and the background dataset size. Dotted entries indicate configurations where the computation time exceeded 1 second, surpassing the real-time threshold. Entries marked with an asterisk (*) denote cases where AD model and data were offloaded to the CPU due to insufficient GPU resources.}    
    \label{fig:timeUsage}
\end{figure}

Figure~\ref{fig:timeUsage} plots the mean execution time per window under the same conditions; dotted entries indicate scenarios in which the processing duration exceeded one second. KernelSHAP failed the real-time requirement in a higher fraction of cases due to the exponential growth of Shapley computations, whereas ShaTS remained comfortably under the limit in most tested settings.

\textbf{Experiment 4: Scalability for larger window sizes}

\begin{figure}
    \centering
    \includegraphics[width=0.8\linewidth]{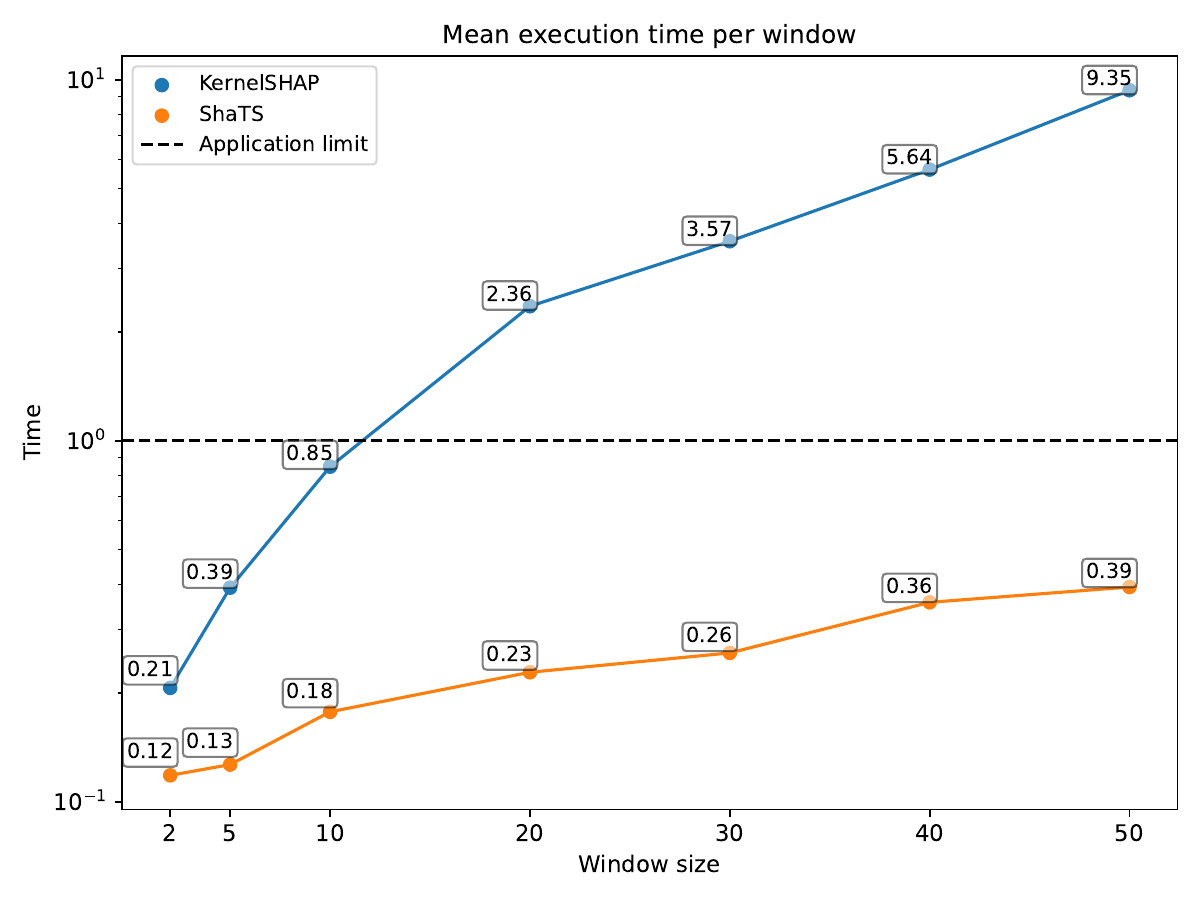}
    \caption{Comparison of mean execution time per window of ShaTS and post hoc SHAP over a batch of 100 windows for different window size AD models}
    \label{fig:windowsAD}
\end{figure}

To demonstrate ShaTS’s efficiency in AD models requiring large time windows, an additional experiment was conducted with 2\,000 coalition subsets and a background dataset size of 40. Although larger windows can capture more extended temporal dependencies and often yield superior performance, Shapley-based xAI methods tend to become increasingly slow as the number of features grows. For this experiment, separate AD models were trained using window sizes of 2, 5, 10, 20, 30, 40, 50, and 60. For 100 windows, SHAP values with post hoc aggregation at the sensor/actuator level and ShaTS values with Sensor/Actuator Grouping were computed, and the average per-window execution time was measured. As shown in Figure~\ref{fig:windowsAD}, post hoc SHAP surpassed the theoretical assumed real-time limit relatively quickly, mainly because the number of Shapley values to compute before aggregating them increased with the window size. In contrast, ShaTS remained nearly constant across all tested window sizes; as it only calculated 44 values (one per sensor/actuator) regardless of the window size, because each group simply expanded in the number of features rather than increasing the total groups. This behavior underscored ShaTS’s scalability, enabling it to handle extended time windows without sacrificing real-time feasibility in industrial applications.

\section{Conclusion and Future Work} \label{sec_conclusion}

This paper introduces ShaTS, an innovative xAI method specifically designed to enhance the understanding of ML/DL models that work with time-series data. Traditional Shapley-based methods often fail to account for the temporal dependencies and inter-feature relationships inherent in time series models, which can lead to less precise and actionable explanations. ShaTS addresses these limitations by applying a priori feature grouping, ensuring that critical relationships are preserved. By offering three grouping strategies (Temporal, Feature, and Multi-Feature), ShaTS provides actionable insights related to different objectives.

The experimental results using the SWaT dataset validate ShaTS’s effectiveness in a practical use case: an AD framework for IIoT. These results demonstrate that ShaTS not only yields temporally consistent explanations but also outperforms conventional SHAP-based approaches that rely on post hoc grouping. Specifically, ShaTS effectively isolates critical instants that mark the transition between anomalous and normal behavior and more reliably identifies the physical sensors/actuators and processes affected by an attack. In contrast, conventional SHAP methods tend to dilute feature contributions when aggregating them after independent computation, making it harder to pinpoint the true sources of anomalies. Moreover, the experiments on resource consumption reveal that ShaTS manages both VRAM consumption and execution time far more efficiently than KernelSHAP. While KernelSHAP frequently exceeds GPU memory limits and fails to meet real-time requirements in many configurations, ShaTS remains computationally feasible even under high-load scenarios. ShaTS scales gracefully when window sizes increase and avoids forcing CPU-based fallback. Consequently, ShaTS stands out as a solid xAI method for real-time IIoT scenarios, offering the dual benefits of precise, actionable explanations and significantly reduced computational overhead.

Future work will focus on exploring ShaTS in other use cases involving ML/DL models that process time-series data. Efforts will also aim to develop advanced Shapley value computation strategies to improve computational efficiency and reduce resource consumption. Additionally, alternative grouping approaches will be investigated to align with the contextual needs of different datasets, further enhancing ShaTS's adaptability and applicability across diverse domains. Furthermore, a comparative study with other SHAP explainers could be carried out to further validate the promising potential and performance of ShaTS.

\begin{table}[ht]
\centering
\begin{scriptsize}
\begin{tabular}{|c||
                >{\centering\arraybackslash}m{1cm}|
                >{\centering\arraybackslash}m{3cm}|
                >{\centering\arraybackslash}m{2cm}||
                >{\centering\arraybackslash}m{1cm}|
                >{\centering\arraybackslash}m{3cm}|
                >{\centering\arraybackslash}m{2cm}|}

\hline
\multirow{2}{*}{\textbf{Attack \#}} & \multicolumn{3}{c|}{\textbf{Attack Point}} & \multicolumn{3}{c|}{\textbf{Attack Process}} \\ \cline{2-7}
                                   & \textbf{Attacked} & \textbf{ShaTS Detected} & \textbf{SHAP Detected} & \textbf{Attacked} & \textbf{ShaTS Detected} & \textbf{SHAP Detected} \\ \hline
\makecell[c]{1} 
& \makecell[c]{MV101}
& \makecell[c]{LIT101 (31.4\%)\\FIT101 (13.8\%)\\PIT502 (10.3\%)} 
& \makecell[c]{FIT101 (16.6\%)\\LIT101 (10.9\%)\\AIT201 (6.2\%)}
& \makecell[c]{P1} 
& \makecell[c]{P1 (60.1\%)\\ P2 (10.3\%)\\ P5 (3.5\%)} 
& \makecell[c]{P1 (33.6\%)\\ P2 (25.6\%)\\ P5 (22.3\%)}\\ \hline

2 
& P102 
& \makecell[c]{P102 (82.1\%)\\ LIT101 (3.7\%)\\PIT502 (2.7\%)}
& \makecell[c]{P102 (11.6\%)\\ AIT402 (7.9\%)\\ MV101 (7.0\%)}
& P1 
&  \makecell[c]{P1 (73.1\%)\\ P3 (8.2\%)\\ P5 (6.1\%)}
& \makecell[c]{P1 (38.4\%)\\ P3 (16.1\%)\\ P5 (15.4\%)}\\ \hline

4 
& MV504 
&  \makecell[c]{PIT502 (13.9\%)\\ AIT501 (8.8\%)\\ AIT201 (8.6\%)}
& \makecell[c]{FIT101 (20.0\%)\\ LIT101 (7.0\%)\\ AIT501 (5.4\%)}
& P5 
&  \makecell[c]{P5 (31.6\%)\\ P2 (16.7\%)\\ P4 (14.8\%)} 
& \makecell[c]{P1 (39.4\%)\\ P2 (23.6\%)\\ P5 (22.5\%)}\\ \hline

6 
& AIT202 
&  \makecell[c]{AIT202 (22.2\%)\\ LIT101 (9.1\%)\\ PIT502 (7.3\%)}
& \makecell[c]{AIT202 (7.7\%)\\ LIT101 (6.3\%)\\ P201 (5.2\%)}
& P2 
&  \makecell[c]{P2 (44.9\%)\\ P3 (21.6\%)\\ P4 (10.2\%)} 
& \makecell[c]{P2 (32.7\%)\\ P3 (17.0\%)\\ P4 (15.6\%)}\\ \hline

7 
& LIT301 
& \makecell[c]{LIT301 (47.7\%)\\ DPIT301 (4.5\%)\\ MV201 (4.5\%)} 
& \makecell[c]{P206 (8.2\%)\\ AIT201 (7.5\%)\\ AIT203 (5.7\%)} 
& P3 
& \makecell[c]{P3 (67.1\%)\\ P5 (15.9\%)\\ P2 (8.3\%)} 
& \makecell[c]{P2 (30.1\%)\\ P3 (27.2\%)\\ P1 (15.8\%)}\\ \hline

8 
& DPIT301 
& \makecell[c]{DPIT301 (51.1\%)\\ MV303 (12.9\%)\\ MV304 (10.4\%)} 
& \makecell[c]{DPIT301 (4.9\%)\\ AIT202 (4.8\%)\\ LIT101 (4.5\%)} 
& P3 
&\makecell[c]{P3 (80.6\%)\\ P5 (14.7\%)\\ P2 (3.6\%)} 
& \makecell[c]{P2 (27.5\%)\\ P5 (23.4\%)\\ P1 (20.1\%)}\\ \hline

10 
& FIT401 
&  \makecell[c]{FIT401 (35.0\%)\\ AIT503 (9.2\%)\\ LIT401 (8.7\%)}
& \makecell[c]{FIT101 (10.4\%)\\ FIT401 (7.0\%)\\ P204 (4.7\%)}
& P4 
&  \makecell[c]{P4 (59.8\%)\\ P3 (24.4\%)\\ P5 (6.8\%)}
& \makecell[c]{P1 (27.1\%)\\ P4 (21.8\%)\\ P5 (19.2\%)}
\\ \hline

17 
& MV303 
&  \makecell[c]{MV301 (40.7\%)\\ FIT301 (8.7\%)\\ P302 (8.7\%)}
& \makecell[c]{P101 (10.1\%)\\ P302 (9.6\%)\\ MV301 (8.7\%)}
& P3 
&  \makecell[c]{P3 (70.8\%)\\ P5 (13.3\%)\\ P6 (8.3\%)}
& \makecell[c]{P3 (39.4\%)\\ P1 (31.0\%)\\ P5 (9.9\%)}
\\ \hline

19 
& AIT504 
&  \makecell[c]{AIT504 (91.7\%)\\ LIT101 (4.1\%)\\ AIT503 (1.8\%)}
& \makecell[c]{AIT201 (17.9\%)\\ AIT504 (17.8\%)\\ LIT301 (17.7\%)}
& P5 
&  \makecell[c]{P5 (82.2\%)\\ P3 (6.4\%)\\ P4 (3.9\%)}
& \makecell[c]{P2 (39.0\%)\\ P5 (38.7\%)\\ P3 (18.7\%)}
\\ \hline

21 
& \makecell{MV101 \\ LIT101} 
&  \makecell[c]{AIT503 (26.7\%)\\ LIT101 (9.8\%)\\ AIT401 (6.7\%)}
& \makecell[c]{FIT101 (13.5\%)\\ FIT201 (6.2\%)\\ AIT202 (6.0\%)}
& P1 
& \makecell[c]{P5 (22.7\%)\\ P1 (15.8\%)\\ P4 (6.9\%)}
& \makecell[c]{P2 (29.6\%)\\ P1 (28.0\%)\\ P5 (23.9\%)}
\\ \hline

22 
& \makecell{UV401 \\ AIT502 \\ P501} 
& \makecell[c]{FIT504 (32.5\%)\\ FIT503 (19.3\%)\\ PIT502 (17.9\%)}
& \makecell[c]{MV201 (5.1\%)\\ FIT101 (5.0\%)\\ FIT401 (4.5\%)}
& \makecell{P4 \\ P5} 
& \makecell[c]{P5 (79.4\%)\\ P1 (4.8\%)\\ P3 (4.8\%)}
& \makecell[c]{P5 (26.2\%)\\ P2 (21.1\%)\\ P4 (20.6\%)}
\\ \hline

23 
& \makecell{P602 \\ DPIT301 \\ MV302} 
&  \makecell[c]{DPIT301 (60.9\%)\\ FIT301 (11.5\%)\\ P302 (8.8\%)}
& \makecell[c]{DPIT301 (6.9\%)\\ AIT201 (5.9\%)\\ AIT202 (5.9\%)}
& \makecell{P3 \\ P6} 
&  \makecell[c]{P3 (84.5\%)\\ P5 (10.4\%)\\ P2 (3.8\%)}
& \makecell[c]{P2 (27.3\%)\\ P1 (22.6\%)\\ P3 (21.5\%)}
\\ \hline

26 
& \makecell{P101 \\ LIT301} 
&  \makecell[c]{P102 (88.0\%)\\ LIT401 (2.3\%)\\ AIT501 (2.3\%)}
& \makecell[c]{P102 (12.4\%)\\ AIT402 (8.3\%)\\ MV302 (6.9\%)}
& \makecell{P1 \\ P3} 
&  \makecell[c]{P1 (82.0\%)\\ P3 (11.8\%)\\ P5 (6.6\%)}
& \makecell[c]{P1 (31.0\%)\\ P3 (19.1\%)\\ P4 (18.0\%)}
\\ \hline

27 
& \makecell{P302 \\ LIT401} 
&  \makecell[c]{LIT401 (15.0\%)\\ LIT301 (9.0\%)\\ AIT203 (4.6\%)}
& \makecell[c]{FIT101 (6.5\%)\\ AIT201 (6.4\%)\\ MV101 (5.3\%)}
& \makecell{P3 \\ P4} 
&  \makecell[c]{P4 (24.8\%)\\ P2 (20.6\%)\\ P3 (15.9\%)}
& \makecell[c]{P2 (30.5\%)\\ P1 (27.3\%)\\ P3 (14.7\%)}
\\ \hline

30 
& \makecell{LIT101 \\ P101 \\ MV201} 
&  \makecell[c]{P102 (91.6\%)\\ P101 (2.8\%)\\ LIT401 (1.6\%)}
& \makecell[c]{P102 (11.3\%)\\ FIT101 (8.7\%)\\ AIT402 (7.1\%)}
& \makecell{P1 \\ P2} 
&  \makecell[c]{P1 (80.8\%)\\ P3 (9.3\%)\\ P5 (6.2\%)}
& \makecell[c]{P3 (21.1\%)\\ P4 (21.0\%)\\ P2 (20.6\%)}
\\ \hline

31
& \makecell{LIT401} 
&  \makecell[c]{LIT401 (23.6\%)\\ FIT101 (12.5\%)\\ P205 (5.4\%)}
& \makecell[c]{FIT101 (7.6\%)\\ MV101 (6.1\%)\\ AIT203 (5.7\%)}
&  \makecell{P4}
&  \makecell[c]{P4 (35.3\%)\\ P2 (27.5\%)\\ P3 (21.9\%)}
& \makecell[c]{P1 (31.2\%)\\ P2 (28.3\%)\\ P3 (16.9\%)}
\\ \hline

32 
& LIT301 
&  \makecell[c]{LIT301 (51.7\%)\\ DPIT301 (4.6\%)\\ P203 (4.4\%)}
& \makecell[c]{P206 (8.9\%)\\ AIT201 (6.9\%)\\ MV304 (6.6\%)}
& P3 
& \makecell[c]{P3 (55.2\%)\\ P2 (17.9\%)\\ P5 (12.1\%)}
& \makecell[c]{P2 (36.1\%)\\ P3 (22.7\%)\\ P1 (15.6\%)}
\\ \hline

34 
& P101 
&  \makecell[c]{P101 (9.1\%)\\ MV101 (8.7\%)\\ LIT401 (6.7\%)}
& \makecell[c]{FIT101 (7.9\%)\\ P102 (4.9\%)\\ AIT402 (4.0\%)}
& P1 
& \makecell[c]{P4 (41.2\%)\\ P2 (16.9\%)\\ P5 (15.6\%)}
& \makecell[c]{P1 (23.3\%)\\ P5 (21.9\%)\\ P3 (20.3\%)}
\\ \hline

36 
& LIT101 
&  \makecell[c]{LIT101 (14.5\%)\\ P203 (10.5\%)\\ P205 (8.0\%)}
& \makecell[c]{FIT101 (9.0\%)\\ LIT101 (8.8\%)\\ FIT301 (7.1\%)}
& P1 
&  \makecell[c]{P1 (29.0\%)\\ P2 (21.5\%)\\ P3 (6.2\%)}
& \makecell[c]{P1 (35.4\%)\\ P2 (20.4\%)\\ P5 (18.2\%)}
\\ \hline

37 
& \makecell{P501 \\ FIT502} 
&  \makecell[c]{FIT504 (31.0\%)\\ FIT503 (24.8\%)\\ PIT502 (15.5\%)}
& \makecell[c]{FIT401 (6.0\%)\\ AIT402 (5.2\%)\\ FIT504 (5.0\%)}
& P5
& \makecell[c]{P5 (60.1\%)\\ P3 (28.5\%)\\ P4 (10.9\%)}
& \makecell[c]{P4 (28.8\%)\\ P3 (18.1\%)\\ P1 (17.8\%)}
\\ \hline

39 
& \makecell{FIT401 \\ AIT502} 
&  \makecell[c]{FIT401 (50.2\%)\\ AIT503 (26.6\%)\\ P201 (6.6\%)}
& \makecell[c]{FIT401 (5.8\%)\\ LIT301 (4.8\%)\\ FIT101 (4.7\%)}
& \makecell{P4 \\ P5} 
&  \makecell[c]{P4 (80.3\%)\\ P5 (12.2\%)\\ P2 (3.4\%)}
& \makecell[c]{P2 (28.1\%)\\ P5 (22.1\%)\\ P4 (17.3\%)}
\\ \hline

\end{tabular}
 \caption{Detailed attribution results for each detected attack in the SWaT dataset. For each attack, the table presents the actual attacked sensor/actuator and process alongside the top-ranked groups detected by ShaTS and conventional SHAP.} \label{table:exp}
\end{scriptsize}
\end{table}

\section{Acknowledgments}

This work has been funded under (a) Grant TED2021-129300B-I00, by MCIN/AEI/10.13039/501100011033, NextGeneration EU/PRTR, UE, (b) Grant PID2021-122466OB-I00, by MCIN/AEI/10.13039/501100011033/FEDER, UE, and (c) the strategic project DEFENDER from the Spanish National Institute of Cybersecurity (INCIBE), by the Recovery, Transformation and Resilience Plan, Next Generation EU.

\clearpage
\bibliographystyle{elsarticle-num}

\end{document}